\documentclass[10pt,twocolumn,letterpaper,pagebackref=true,breaklinks=true,letterpaper=true,colorlinks,bookmarks=false]{article}

\usepackage{enumitem} 

\usepackage{wacv}
\usepackage{times}
\usepackage{epsfig}
\usepackage{graphicx}
\usepackage{amsmath}
\usepackage{amssymb}
\usepackage{hyperref}
\usepackage{subcaption}
\usepackage{booktabs}
\usepackage{tabularx}
\usepackage[T1]{fontenc}  %

\usepackage{microtype}

\usepackage{balance}
\usepackage[sort, numbers]{natbib}
\usepackage[dvipsnames]{xcolor}
\usepackage{enumitem}

\wacvfinalcopy %

\def\wacvPaperID{89} %

\ifwacvfinal\fi
\def \AnnotationTaskPayment {\$0.8}

\def \AnnotationTaskFilterSD {40\%}
\def \AnnotationTaskFilterPCT {60\%}

\def \UserExperimentUserN {12}
\def \UserExperimentImageCandidateN {600}
\def \UserExperimentImagePCT {5\%}
\def \UserExperimentParticipantN {333} 
\makeatletter
\long\def\comment#1{}
\long\def\@makecaption#1#2{
   \setbox\@tempboxa\hbox{\small \noindent {\small \bf #1.}~\small#2}
   \setlength{\@ctmp}{\hsize}
   \addtolength{\@ctmp}{-\@figindent}\addtolength{\@ctmp}{-\@figindent}
   \ifdim \wd\@tempboxa >\@ctmp
      {\small {\small\bf #1.}~\small#2\par}
   \else
      \hbox to\hsize{\hfil\box\@tempboxa\hfil}
  \fi\vspace*{\belowcaptionskip}}

\renewcommand\section{\@startsection {section}{1}{\z@}%
                                   {-.8ex \@plus -2ex \@minus -.2ex}%
                                   {.5ex \@plus.2ex}%
                                   {\normalfont\large\bfseries\raggedright}}
\renewcommand\subsection{\@startsection{subsection}{2}{\z@}%
                                   {-.6ex \@plus -2ex \@minus -.2ex}%
                                   {.5ex \@plus.2ex}%
{\normalfont\large\bfseries\raggedright}}
\renewcommand\subsubsection{\@startsection{subsubsection}{3}{\z@}%
                                     {-.5ex\@plus -.2ex \@minus -.2ex}%
                                     {.2ex \@plus .2ex}%
                                     {\normalfont\large\bfseries\raggedright}}

  \renewenvironment{thebibliography}[1]{%
    \begin{oldthebibliography}{#1}%
      \small
      \setlength{\parskip}{0pt}%
      \setlength{\itemsep}{0pt}%
  }%
  {%
    \end{oldthebibliography}%
  }

\renewcommand\normalsize{%
   \@setfontsize\normalsize\@xpt\@xiipt
   \abovedisplayskip 4\p@ \@plus2\p@ \@minus5\p@
   \abovedisplayshortskip \z@ \@plus3\p@
   \belowdisplayshortskip 2\p@ \@plus3\p@ \@minus3\p@
   \belowdisplayskip \abovedisplayskip
   \let\@listi\@listI}
\renewcommand\small{%
   \@setfontsize\small\@ixpt{11}%
   \abovedisplayskip 2.5\p@ \@plus3\p@ \@minus4\p@
   \abovedisplayshortskip \z@ \@plus2\p@
   \belowdisplayshortskip 2\p@ \@plus2\p@ \@minus2\p@
   \def\@listi{\leftmargin\leftmargini
               \topsep 4\p@ \@plus2\p@ \@minus2\p@
               \parsep 2\p@ \@plus\p@ \@minus\p@
               \itemsep \parsep}%
   \belowdisplayskip \abovedisplayskip
}

\def\tightmath{
\abovedisplayskip=3pt plus 2pt minus 1pt
\abovedisplayshortskip=0pt plus 1pt minus 1pt
\belowdisplayskip=3pt plus 2pt minus 1pt
\belowdisplayshortskip=0pt plus 1pt minus 1pt }
\def\crushmath{
\abovedisplayskip=1pt plus 1pt minus 2pt
\abovedisplayshortskip=1pt plus 1pt minus 2pt
\belowdisplayskip=1pt plus 1pt minus 2pt
\belowdisplayshortskip=1pt plus 1pt minus 2pt }
\tightmath
\crushmath

\makeatother

\begin{document}

\title{Understanding Image Quality and Trust in Peer-to-Peer Marketplaces}

\author{
Xiao Ma$^1$,
Lina Mezghani$^{2\dagger}$,
Kimberly Wilber$^{1,3\dagger}$,
\\
Hui Hong$^4$,
Robinson Piramuthu$^4$,
Mor Naaman$^1$,
Serge Belongie$^1$
\\
{\tt\small \{xm75,mjw285,mor.naaman,sjb344\}@cornell.edu; lina.mezghani@polytechnique.edu;}
\\
{\tt\small \{huihong, rpiramuthu\}@ebay.com}
\\
$^1$ Cornell Tech; $^2$ \'Ecole polytechnique; $^3$ Google Research; $^4$ eBay, Inc.
\vspace{-10pt}
}

\newcommand\blfootnote[1]{%
  \begingroup
  \renewcommand\thefootnote{}\footnote{#1}%
  \addtocounter{footnote}{-1}%
  \endgroup
}

\maketitle
\ifwacvfinal\thispagestyle{empty}\fi

\begin{abstract}

As any savvy online shopper knows, second-hand peer-to-peer marketplaces are filled with images of mixed quality.
How does image quality impact marketplace outcomes, and can quality be automatically predicted?
In this work, we conducted a large-scale study on the quality of user-generated images in peer-to-peer marketplaces.
By gathering a dataset of common second-hand products ($\approx$75,000 images) and annotating a subset with human-labeled quality judgments, we were able to model and predict image quality with decent accuracy ($\approx$87\%).
We then conducted two studies focused on understanding the relationship between these image quality scores and two marketplace outcomes: sales and perceived trustworthiness.
We show that image quality is associated with higher likelihood that an item will be sold, though other factors such as view count were better predictors of sales.
Nonetheless, we show that high quality user-generated images selected by our models outperform stock imagery in eliciting perceptions of trust from users. 
Our findings can inform the design of future marketplaces and guide potential sellers to take better product images.

\end{abstract} 
\vspace{-10pt}
\section{Introduction}
\vspace{-10pt}

\blfootnote{$^\dagger$ Work done while at Cornell Tech.}

Product photos play an important role in online marketplaces.
Since it is impractical for a customer to inspect an item before purchasing it, photos are especially important for decision making in online transactions.
Good photos reduce uncertainty and information asymmetry inherent in online marketplaces~\cite{akerlof1978market}.
For example, good images help potential customers understand whether products are likely to meet their expectations, and on peer-to-peer marketplaces, an honest photo can be an indication of wear and tear. 
However, in peer-to-peer marketplaces where buyers and sellers are very often non-professionals --- without specialized photography expertise nor access to high-end studio equipment --- 
it is challenging to know how to take high-quality pictures.

\begin{figure}[t!]
  \centering
\includegraphics[width=1\linewidth]{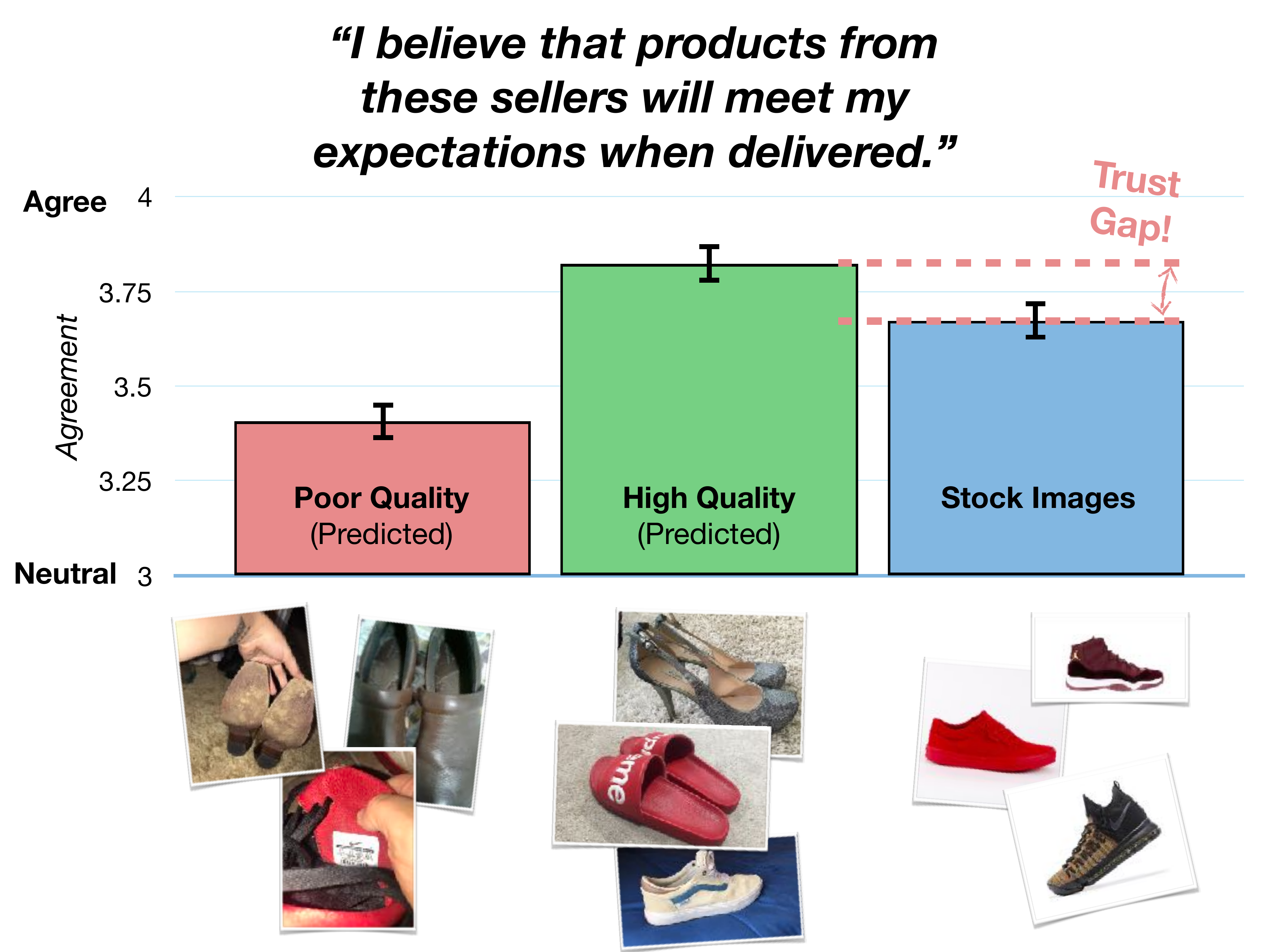}
\caption{\label{fig:trust-gap}
  \textbf{We study the interplay between image quality, marketplace outcomes, and user trust in peer-to-peer online marketplaces.} Here, we show how image quality (as measured by a deep-learned CNN model) correlates with user trust. User studies in Sec.~\ref{sec:user-trustworthiness} show that high quality images selected by our model out-performs stock-imagery in eliciting user trust.
}
\end{figure}

The problem of having images with diverse range of qualities is becoming more urgent as mobile marketplaces like LetGo.com, Facebook Marketplace, etc.\ become more popular.
These applications facilitate easy listing: users can snap a picture with their phone, type in a short description, and instantly attract customers.
However, it is difficult and time-consuming to take nice photos.
If the marketplace is saturated with unappealing amateur product photos, users will lose trust in their shopping experience, negatively impacting sellers as well as the reputation of the marketplace itself.
In this work, we present a computational approach for modeling, predicting, and understanding image quality in online marketplaces.

\textbf{Modeling and predicting image quality.} 
Previous work studied general photo aesthetics, but we show existing models do not completely capture image quality in the context of online marketplaces.
To do this, we used both black-box neural networks and interpretable regression techniques to model and predict \emph{image quality} (how appealing the product image appears to customers).
For the regression-based approach, we model the factors of the photographic environment that influence quality with handcrafted features, which can guide the potential sellers to take better pictures.
As part of our modeling process, we built and curated a publicly available dataset of images from LetGo.com, annotated with image quality scores for two product categories, \emph{shoes} and \emph{handbags}.
We showed that our models outperform baseline models trained only to capture aesthetic judgments.

\textbf{Marketplace outcomes: sales and perceived trustworthiness.} Using our learned quality model, we then showed that image quality scores are associated with two different group outcomes, sales and perceived trustworthiness.
Predicted image quality was associated with higher likelihood that an item is sold, while high quality user-generated images selected by our model out-performs stock imagery in eliciting perceived trust from users (see~\autoref{fig:trust-gap}).

In short, here are the contributions of the paper: (1) We curated a dataset of common second-hand products ($\approx$75,000 images) and annotated a third of them with human-labeled quality judgment; 
(2) We developed a better understanding of how visual features impact image quality,
while training a convolutional network to predict image quality with decent accuracy ($\approx$87\%).
(3) We showed that predicted image quality is associated with positive marketplace outcomes such as sales and perceived trust.
Our findings can be valuable for designing better online marketplaces, or to help sellers with less photographic experience
take better product images.

\begin{figure}[t]
\centering
\includegraphics[width=0.85\linewidth]{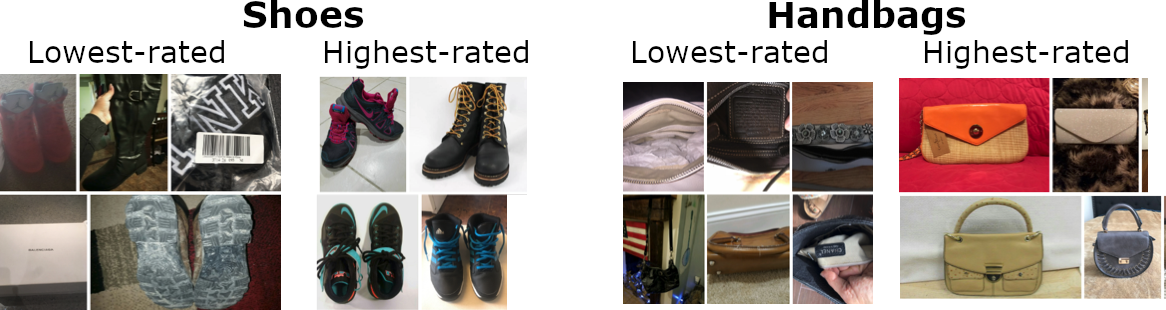}
\caption{\label{fig:example-lohi} Samples of lowest-rated and highest-rated images from \emph{shoe} and \emph{handbag} groundtruth.}
\end{figure}

\section{Related Work}

\textbf{Image Quality in Online Marketplaces.} Previous work has shown that image-based features such as brightness, contrast, and background lightness contribute to the prediction of click-through rate (CTR) in the context of product search~\cite{goswami2011study,chung2012impact}.
This line of work used hand-crafted image features, but did not actually assess the image quality as dependent variable.
In another work on eCommerce, image quality was modelled and predicted through linear regression, and shown to be significant predictors of buyer interest~\cite{di2014picture}. However, the dataset is not available, nor any details on the modelling methodology or model performance.
Our work fills this gap by contributing a large annotated image quality dataset, along with improved model performance.

\textbf{Automatic Assessment of Aesthetics.} Another line of closely related work is automatically assessing the aesthetic quality of images.
Early work on aesthetic quality frequently used handcrafted features, for example~\cite{datta2006studying,li2010towards,bhattacharya2010framework}. More recent work focused on AVA, a large scale dataset containing 250,000 images with aesthetic ratings~\cite{murray2012ava} and adapting deep learning features~\cite{lu2014rapid} to improve prediction accuracy. 
In addition, aesthetics quality has been shown to predict the performance of images in the context of photography, increasing the likelihood that a picture will receive ``likes'' or ``favorites''~\cite{schwarz2016will}.
Although aesthetic quality is a fundamental and important feature of imagery, images online marketplaces belong to another visual domain compared to most of the images in this line of work.
We show that aesthetic quality models do not completely capture product image quality.

\textbf{Web Aesthetics and Trust.} A large body of work in human computer interaction has investigated the link between the marketplace website's aesthetics and perceived credibility. For example, previous work has shown low level statistics such as balance and symmetry correlate with aesthetics~\cite{zheng2009correlating,michailidou2008visual}. More recently, computational models have been developed to capture visual complexity and colorfulness of website screenshots to predict aesthetic appeal~\cite{reinecke2013predicting}.
However, since product images take up majority of the space in online marketplaces, inadequate attention has been paid to how the image quality, rather than interface quality, impact user trust.
Our work focuses on product images as the most salient visual element of online marketplaces to study how they contribute to user trust.

Our work is concerned with \emph{user trust}, which cannot be measured from purchase outcomes and click-through-rates.
Perceptions of user trust is important for several reasons: trust influences loyalty, purchase intentions~\cite{hong2011impact}, retention~\cite{sun2010sellers}, and is important for the platform's initial adoption and growth~\cite{su10010291}.
This is why we take a ``social science'' approach when soliciting trust judgments in Sec.~\ref{sec:user-trustworthiness}.

\textbf{Domain Adaption: Matching street photos to stock photos.} One might wonder why marketplaces do not simply retrieve the stock photo of the product being depicted and display it instead. There are some problems with this approach. First, in used goods markets, stock photos do not depict the actual item being sold and are generally discouraged~\cite{bland2007risk}. Second, stock image retrieval is a computationally challenging task, with state-of-the-art methods~\cite{wang2016matching} achieving around 40\% top-20 retrieval accuracy on the Street2Shop~\cite{kiapour2016buy} dataset. Our work also contributes to the dataset of ``street'' photos taken by different users using a variety of mobile devices in varying lighting conditions and backgrounds. %
\section{Datasets}
As shown in previous work~\cite{di2014picture}, image quality matters more for product categories that are inherently more visual (e.g., clothing).
Thus in our development of the dataset, we focus on the \emph{shoe} and \emph{handbag} categories. These two categories are among the most popular goods found on secondhand marketplaces and are visually distinctive enough to pose an interesting computer vision challenge.

There are two sources for the data used in this work: LetGo.com, and eBay. We focused on the publicly available LetGo.com images for creating the hand-annotation dataset, and used private data from eBay to test the relationship between image quality and marketplace outcome -- sales.

\subsection{LetGo.com}
LetGo.com is a mobile-based peer-to-peer marketplace for used items, similar to Craiglist. Potential buyers can browse through the ``listings'' made by sellers and contact the seller to complete the transaction out of the platform.

We collected product images data for two product categories, shoes and handbags. We crawled the front page of LetGo.com every ten to thirty minutes for a month, filtering the listings by relevant keywords in the product listing caption.
For \emph{shoes}, we used the keywords ``shoe,'' ``sandal,'' ``boot,'' or ``sneaker'' and collected data between November to December 2017 (66,752 listings containing 133,783 images in total).
For \emph{handbags}, we used the keywords ``purse'' or ``handbag' and collected data between April to May 2018 (29,839 listings containing 44,725 images in total).

\subsection{eBay}\label{sec:online-marketplace}
To understand whether image quality impacts real world outcomes, we partnered with eBay, one of the largest online marketplaces.

We collected data for listings on eBay in our two product categories, shoes and handbags, including the product images, meta-data associated with the listing, as well as whether the listing had at least one sales completed before becoming expired. 
We sampled data based on the date on which the listing expired (during May 2018).
We also down-sampled the available listings to create a balanced set of sold and unsold listings.
In summary, this dataset included 66,000 sold and 66,000 unsold listings for \emph{shoes}, and 16,000 sold and 16,000 unsold listings for \emph{handbag}.

To evaluate our model's generalizability in a real-world setting, we train models on publicly available LetGo.com data and test the relationship between predicted image quality and sales on eBay. 
\section{Annotating Image Quality}\label{sec:annotation}
We collected ground truth image quality labels for our LetGo.com dataset (LetGo below) using a crowdsourcing approach.
We designed the following task and issued it on Amazon Mechanical Turk, paying \AnnotationTaskPayment~per task.
Each task contained 50 LetGo images randomly batched, and at least 3 workers rated each image.
For each image, the worker was asked to rate the image quality on a scale between 1 (not appealing) to 5 (appealing).
In total, we annotated 12,515 images from the \emph{shoe} category and 12,222 from the \emph{handbag} category. Only LetGo data was used for this task.

An important consideration is the difference between \emph{product} quality and \emph{photographic} quality. In this survey, we are primarily interested in what the merchant can do to make their listings more appealing, so it is important that workers ignore perceived product differences. To help prime our workers along this line of thinking, we added two text survey questions spaced throughout each task with the following prompt:
\emph{
``Suppose your friend is using this photo to sell their shoes. What advice would you give them to make it a better picture?
How can the seller improve this photo?''
}
This task also slows workers down and forces them to carefully consider their choices.

After collecting the data, we first standardized each worker's score distribution to zero mean and unit variance to account for task subjectivity and individual rater preference.
Then, we filtered out images where the standard deviation of all rater scores was within the top \AnnotationTaskFilterSD, retaining \AnnotationTaskFilterPCT~of the original dataset. This removed images where annotators strongly disagree. This filtering also improved the inter-rater agreement as measured by an average pairwise Pearson's $\rho$ correlation across each rater from 0.34 $\pm$ 0.0046 on the unfiltered shoe data to 0.70 $\pm$ 0.0031 on the filtered shoe data.
This shows our labeling mechanism can reliably filter images with low annotation agreement.

\begin{figure}[t]
\centering
\includegraphics[width=0.45\linewidth]{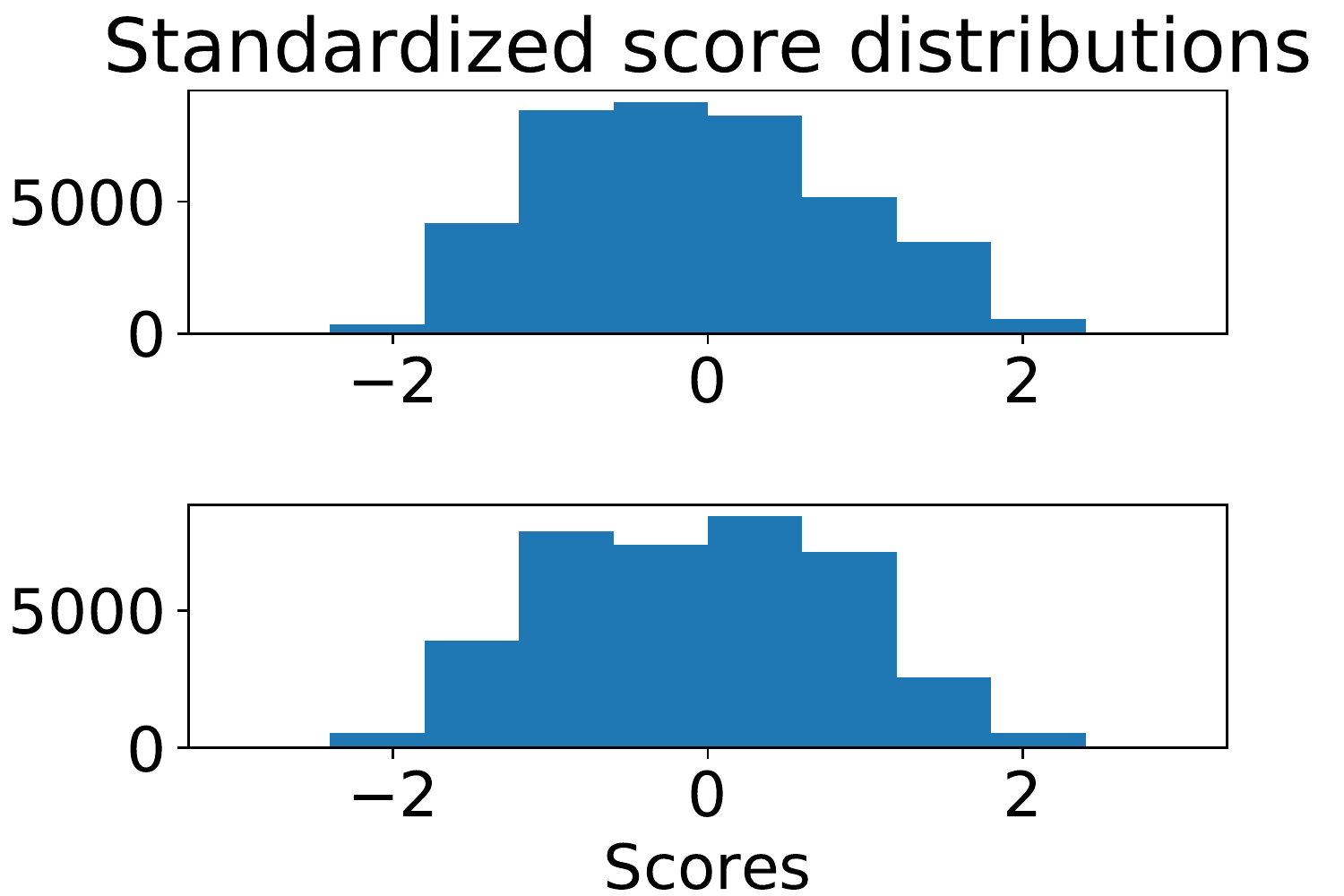}
\includegraphics[width=0.45\linewidth]{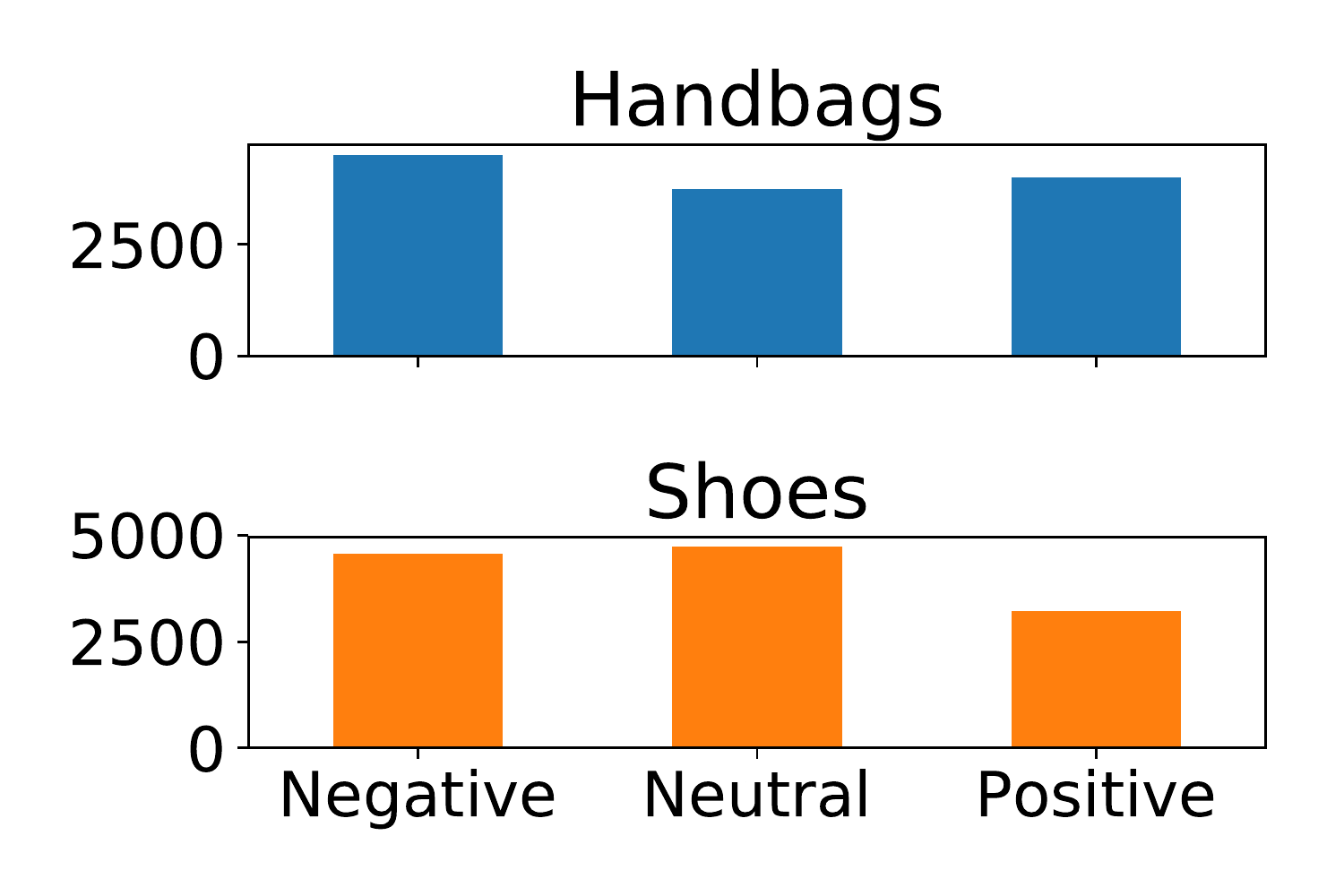}
\caption{\label{fig:datasets}\emph{Left:} Standardized score distributions for filtered images on \emph{shoe} and \emph{handbag} categories. \emph{Right:} Final ground truth labels, showing a fairly even distribution.
}
\end{figure}

Finally, we discretized the scores into three image quality labels: good, neutral and bad. To do this, we rounded the average score to the nearest integer, and took the positive ones as good, zeros as neutral, and negative ones as bad. The resulting \emph{shoe} and \emph{handbag} datasets are roughly balanced; see score distributions in~\autoref{fig:datasets} and example images in~\autoref{fig:example-lohi}.

\section{Modeling Image Quality}
After collecting ground truth for our dataset, we study what factors of an image can influence perceived quality. This turns out to be nontrivial. Shopping behavior is complicated, and customers and crowd workers alike may have intricate preferences, behaviors, and constraints.

We attempt to model image quality in two ways: first, we train a deep-learned CNN to predict ground truth labels. This model is fairly accurate, allowing us to approximate image quality on the rest of our dataset, but as a ``black box'' model it is largely uninterpretable, meaning it does not reveal what high-level image factors lead to high image quality.
Second, to understand quality at an interpretable level, we then use multiple ordered logistic regression to predict the dense quality scores. This lets us draw conclusions about what photographic aspects lead to perceived image quality.

\textbf{Prediction}
For our model, we use the pretrained \emph{Inception v3} network architecture~\cite{szegedy2016rethinking} provided by PyTorch~\cite{paszke2017automatic}.
Our task is 3-way classification: given a product image, the model predicts one of $\{0, 1, 2\}$ for Negative, Neutral, and Positive respectively. To do this, we remove the last fully-connected layer and replace it with a linear map down to 3 output dimensions.

Image quality measurements are subjective. Even though our crowdsourcing worker normalization filters data where workers disagree, we want to allow the model to learn some notion of uncertainty. To do this, we train using \emph{label smoothing}, described in \S~7.5 of \cite{Goodfellow-et-al-2016}. We modify the negative log likelihood loss as follows. Given a single input image, let $x_i$ for $i \in \{0, 1, 2\}$ be the three raw output scores (logits) from the model and let $\hat{x}_i = \log{\exp{x_i} / \sum_j \exp{x_j}}$ be the output log-probabilities after softmax. To predict class $c \in \{0,1,2\}$, we use a modified label \emph{smoothing loss},
$ \ell(x, c) = - (1-\epsilon) \hat{x}_c - \epsilon \sum_{i \neq c}  \hat{x}_i $
for some smoothing parameter $\epsilon$, usually set to $0.05$ in our experiments. This modified loss function avoids penalizing the network too much for incorrect classifications that are overly confident.

These models were fine-tuned on our LetGo dataset for 20 epochs  (\emph{shoe}: $N$=12,515; \emph{handbag}: $N$=12,222). We opted not to use a learning rate schedule due to the small amount of data.

\textbf{Aesthetic Quality Baseline}
We also considered a baseline aesthetic quality prediction task to test whether existing models that capture photographic aesthetics can generalize to predict product image quality. We fine-tuned an Inception~v3 network on the AVA Dataset~\cite{murray2012ava}. To transform predictions into outputs suited to our dataset, we binned the mean aesthetic annotation score from AVA into positive, neutral, and negative labels. The model was fine-tuned on AVA for 5 epochs.

\textbf{Evaluation} We used a binary ``forced-choice'' accuracy metric: starting from a held-out set of positive and negative examples, we considered the network output to be positive if $x_2$>$x_0$, effectively forcing the network to decide whether the positive elements outweigh the negative ones, removing the neutral output. By this metric, our best shoe model achieved 84.34\% accuracy and our best handbag model achieved 89.53\%. This indicates that our model has a reasonable chance of agreeing with crowd workers about the overall quality sentiment of a listing image.\footnote{
If we include neutral images and predictions and simply compare the 3-way predicted output to the ground truth label on the entire evaluation set, our handbag model achieves 64.09\% top-1 accuracy and our shoe model achieves 58.36\%. %
}
For comparison, the baseline aesthetic model achieved 68.8\% accuracy on shoe images and 78.8\% accuracy on handbags.
This shows that product image quality cannot necessarily be predicted by aesthetic judgments alone, and that our dataset constitutes a unique source of data for the study of online marketplace images.
Our image quality model gives us a black box understanding of image quality in an uninterpretable way. 
Next, we investigate what factors influence image quality. %
\subsection{What Makes a Good Product Photo?}

Guiding sellers to upload good photos for their listings is a challenge that almost all online eCommerce sites face. Many sites provide photo tips or guidelines for sellers. For example, Google Merchant Center\footnote{\url{https://support.google.com/merchants/answer/6324350?hl=en&ref_topic=6324338}}
 suggests to ``use a solid white or transparent background'', or to ``use an image that shows a clear view of the main product being sold''. eBay also provides ``tips for taking photos that sell''\footnote{\url{https://pages.ebay.com/seller-center/listing-and-marketing/photo-tips.html}}, including ``tip \#1: use a plain, uncluttered backdrop to make your items stand out'', ``tip \#2: turn off the flash and use soft, diffused lighting''. In addition, many online blogs and YouTube channels also provide tutorials on how to take better product photos.

Despite the abundance of product photography tips, few previous work has validated the effectiveness of these strategies computationally (with the exception of~\cite{chung2012impact,goswami2011study}). Although there is a robust line of research on computational photo aesthetics (e.g.,~\cite{schwarz2016will}), product photography differs greatly in content and functionality from other types of photography and is worth special examination.

In this work, we leverage our annotated dataset, and conducted the first computational analysis of the impact of common product photography tips on image quality. Unlike previous work~\cite{chung2012impact,goswami2011study} that analyzed the impact of image features on clicks, we evaluate directly on potential buyers' perception of image quality. In a later section, we then show how image quality can in turn predict sales outcomes.

\subsubsection{Selecting Image Features}
In order to select the image features to validate, we took a qualitative approach and analyzed 49 online product photography tutorials. We collected the tutorials through Google search queries such as ``product photography'', ``how to take shoe photo sell'', and took results from top two pages (filtering out ads). 
We manually read and labeled the topics mentioned in these tutorials and summarized the most frequently mentioned tips. Out of the 49 tutorials we analyzed, the most frequent topics were: (1) Background (mentioned in 57\% of the tutorials): keywords included white, clean, uncluttered; (2) Lighting (57\%): soft, good, bright; (3) Angles (40\%): multiple angles, front, back, top, bottom, details; (4) Context (29\%): in use; (5) Focus (22\%): sharp, high resolution; (6) Post-Production (22\%): white balance, lighting, exposure; and (7) Crop (14\%): zoom, scale.

Based on the qualitative results, as well as referencing previous work~\cite{chung2012impact,goswami2011study}, we defined and calculated a set of image features to analyze for their impact on image quality. There are three types of image features that we considered: (1) Global features such as brightness, contrast, and dynamic range; (2) Object features based on our object detector (more details on that below); and (3) Regional features focusing on background and foreground. ~\autoref{tab:feature_table} contains the definition and example images for our complete set of image features.

\begin{table}[]
{\small
\begin{tabular}{m{.3\linewidth} m{.35\linewidth} m{.08\linewidth} m{.08\linewidth}}
\toprule
\textbf{Feature Name}
& \textbf{Definition}
& \textbf{Low}
& \textbf{High} \\
\midrule

\multicolumn{4}{l}{\textbf{\textit{Global Features:}}}  \\

brightness
& 0.3R + 0.6G + 0.1B
& \includegraphics[width=\linewidth]{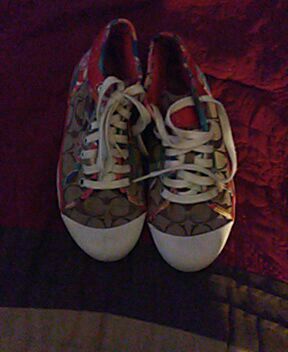}
& \includegraphics[width=\linewidth]{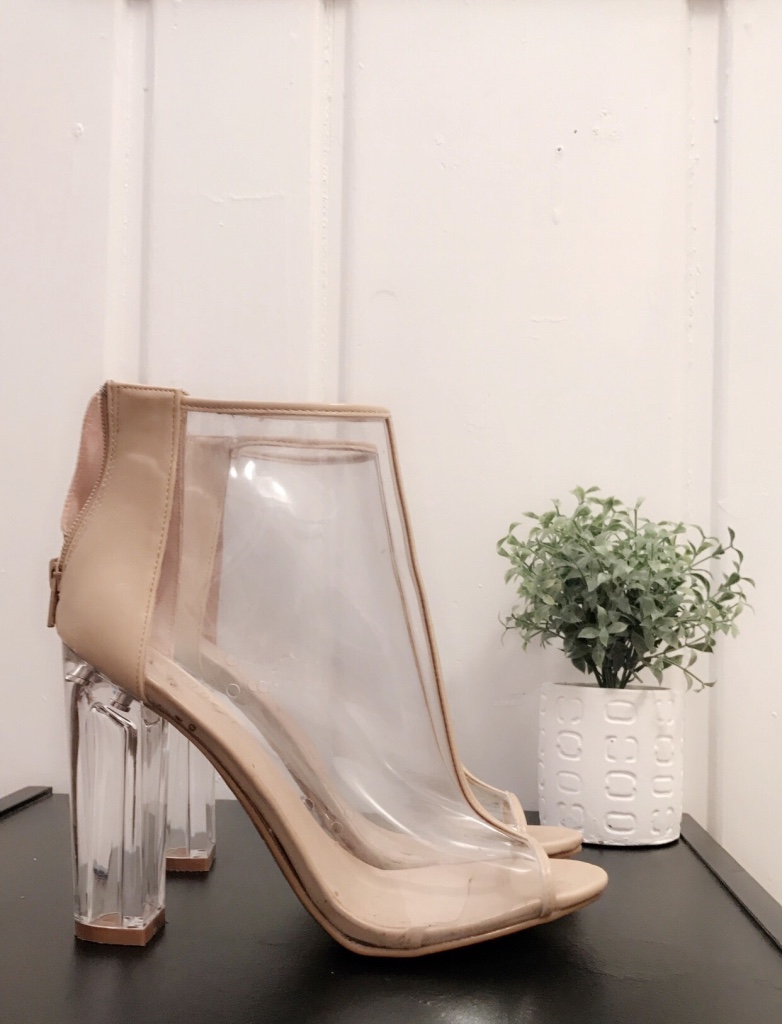}
\\

contrast
& Michelson contrast
& \includegraphics[width=\linewidth]{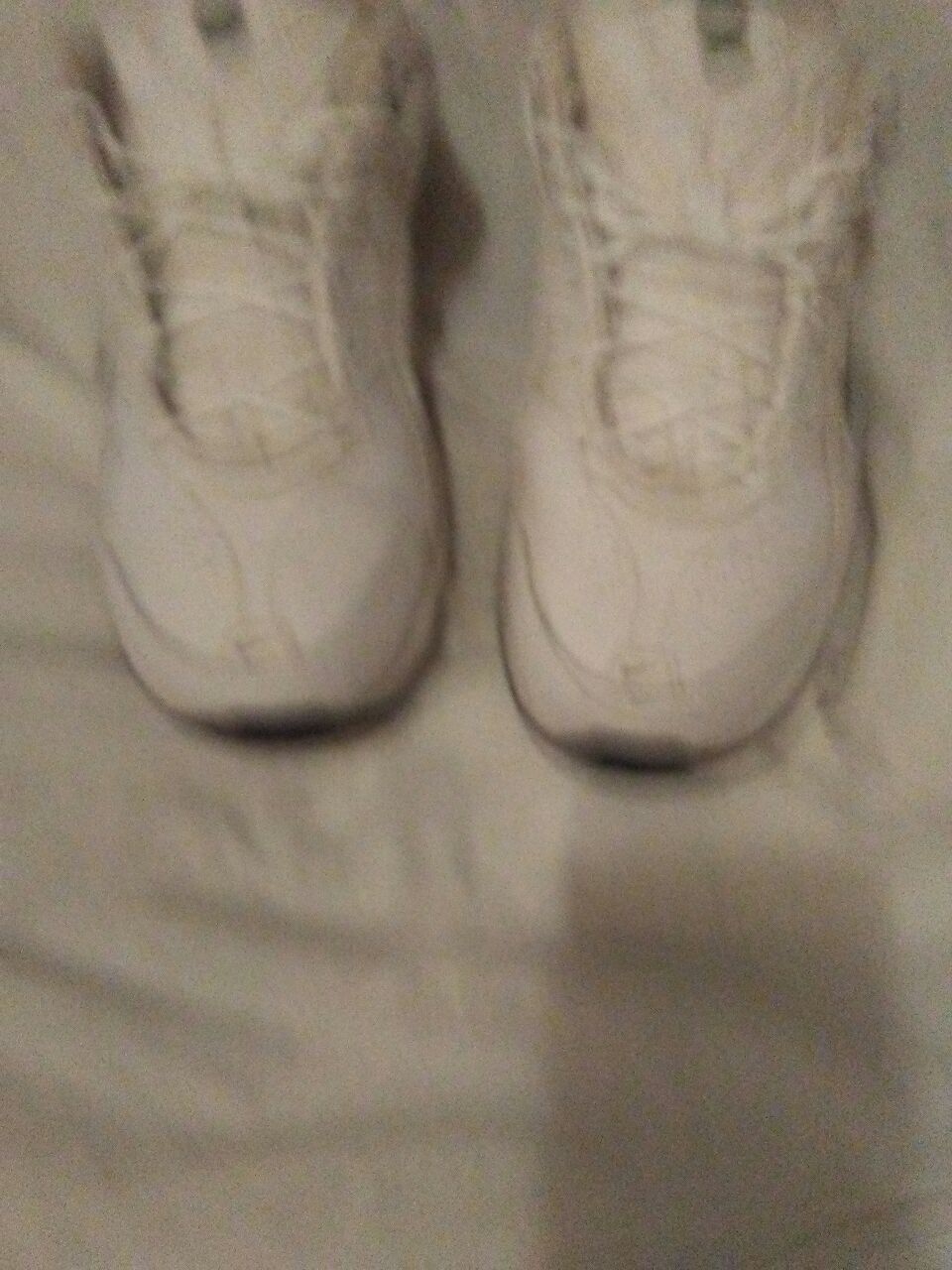}
& \includegraphics[width=\linewidth]{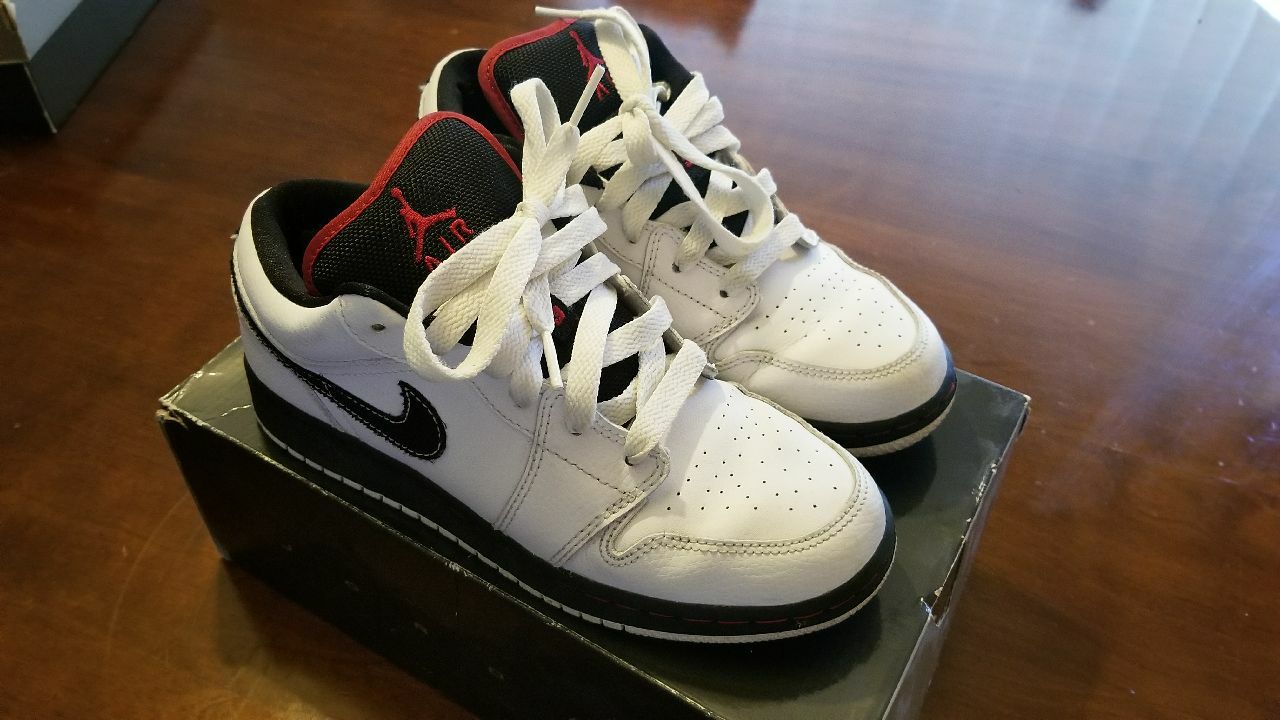}
\\

dynamic\_range
& grayscale (max - min)
& \includegraphics[width=\linewidth]{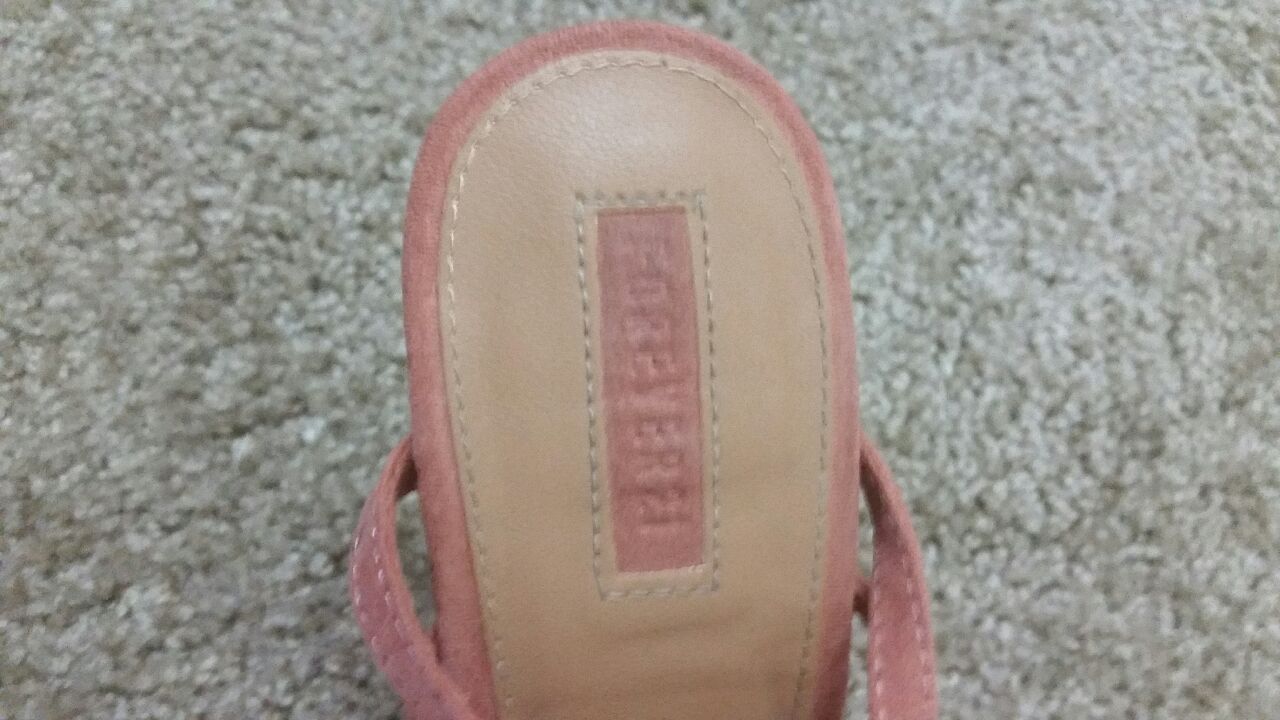}
& \includegraphics[width=\linewidth]{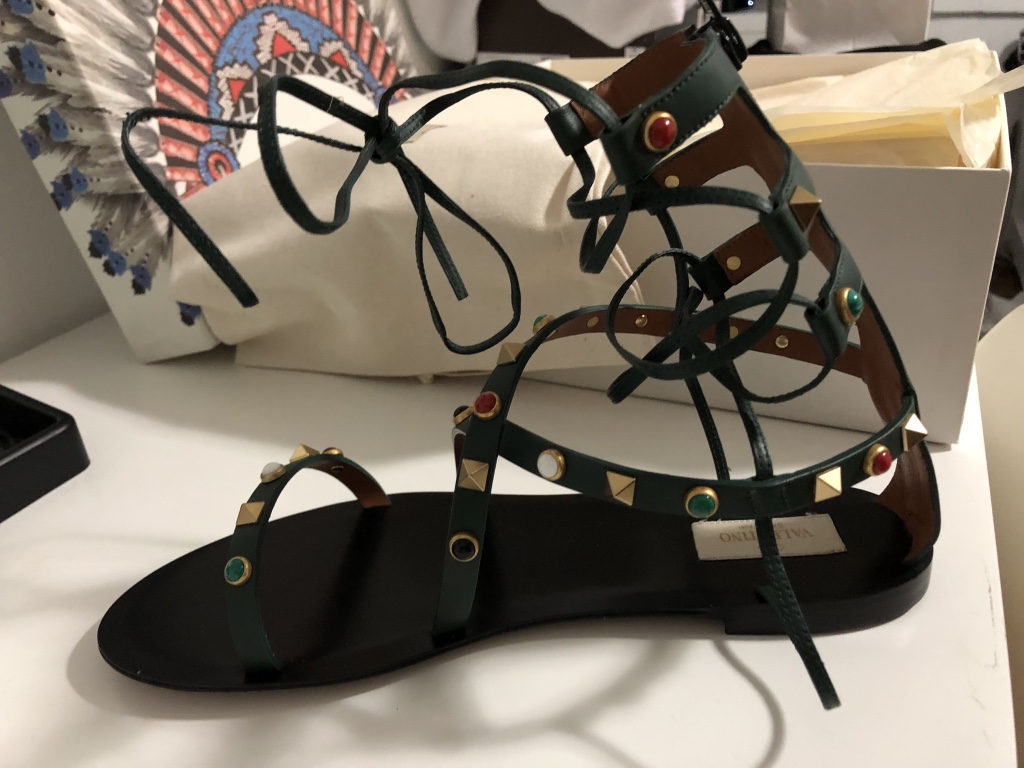}
\\

width
& \multicolumn{3}{l}{the width of the photo in px}
\\

height
& \multicolumn{3}{l}{the height of the photo in px}
\\

resolution
& \multicolumn{3}{l}{width * height / $10^6$}
\\

\hline
\multicolumn{4}{l}{\textbf{\textit{Object Features:}}}  \\
object\_cnt
& \# of objects detected
& \includegraphics[width=\linewidth]{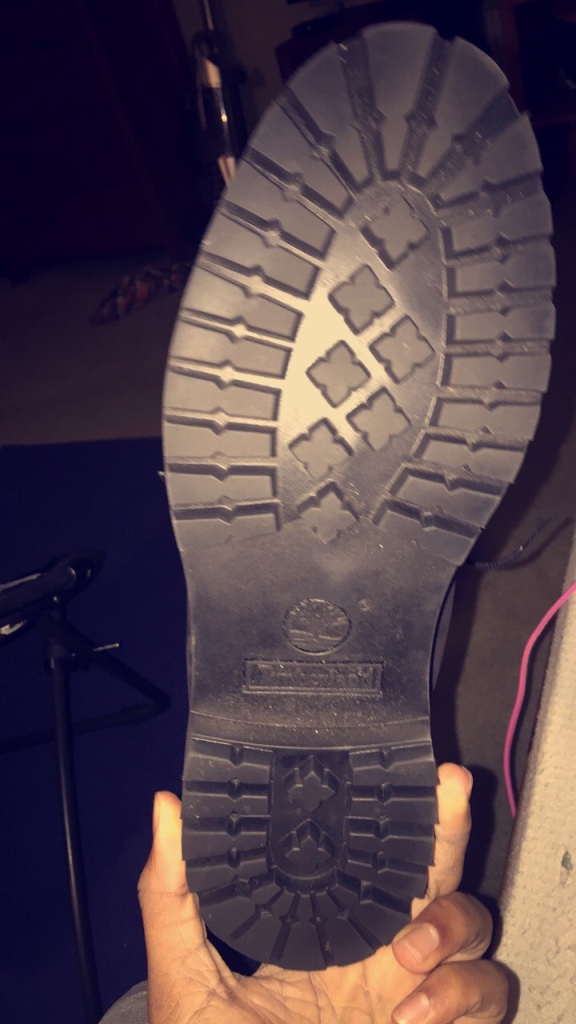}
& \includegraphics[width=\linewidth]{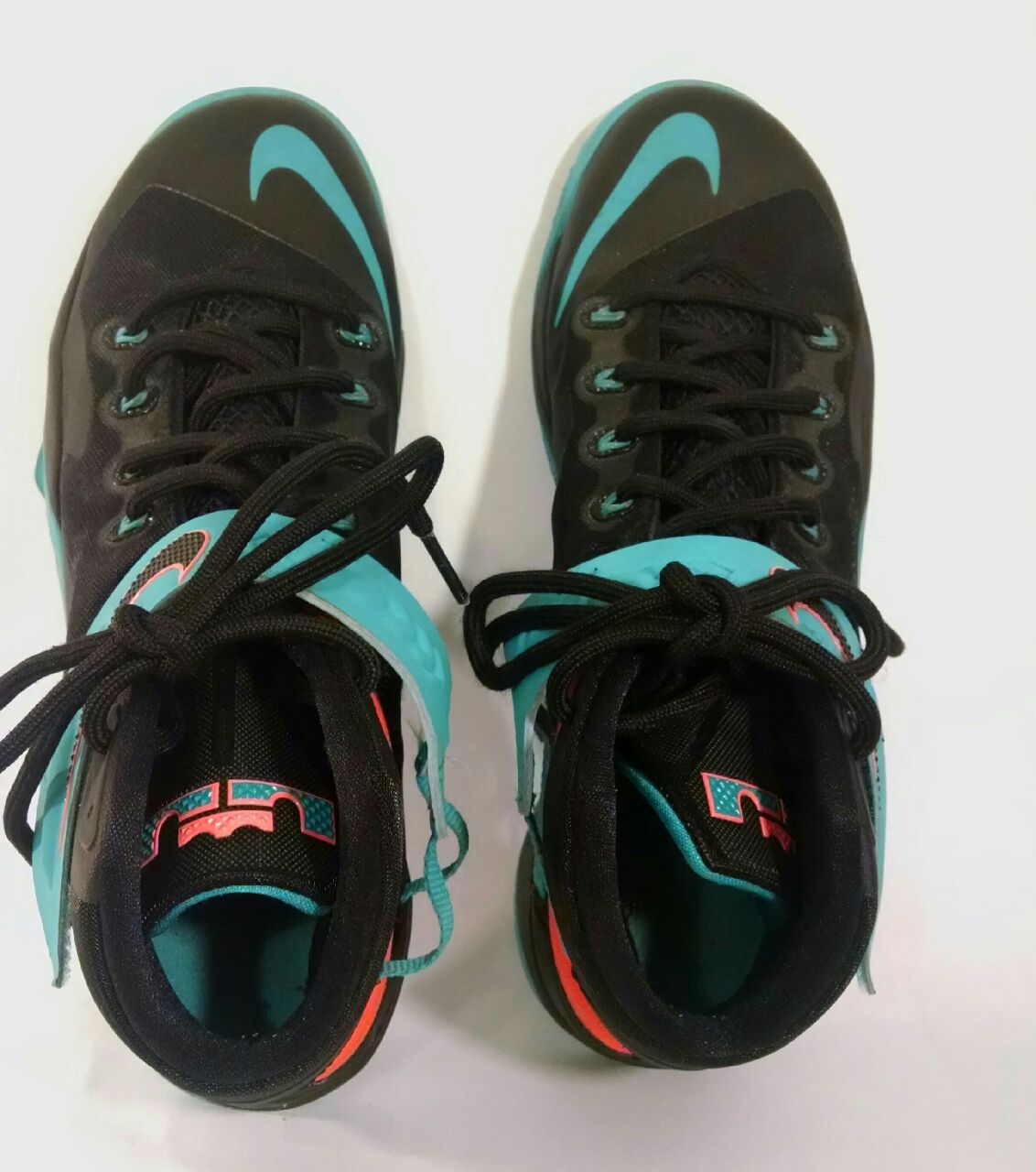}
\\

top\_space
& bounding box top to top of image in px
& \includegraphics[width=\linewidth]{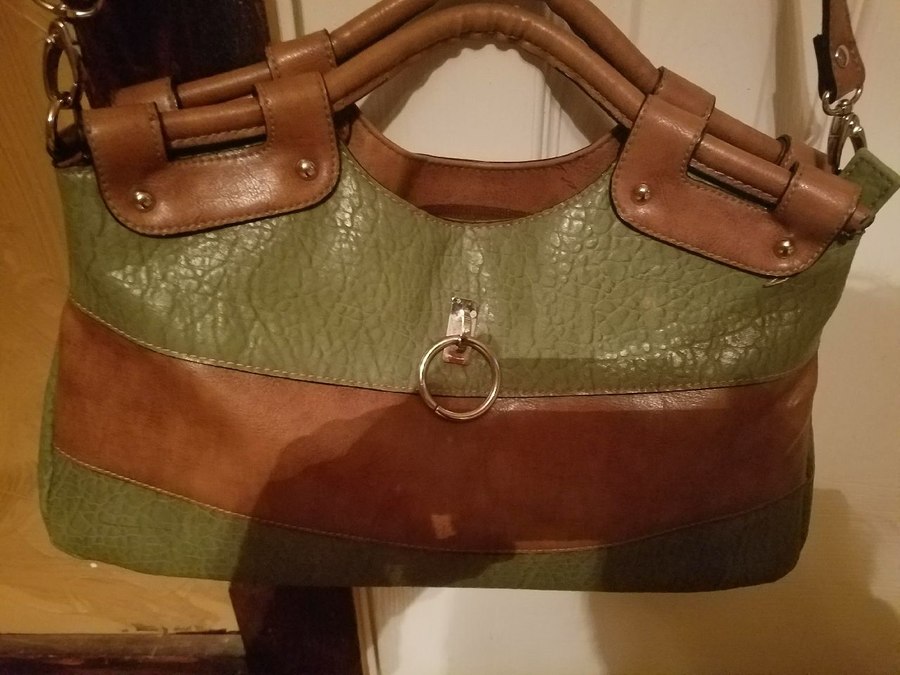}
& \includegraphics[width=\linewidth]{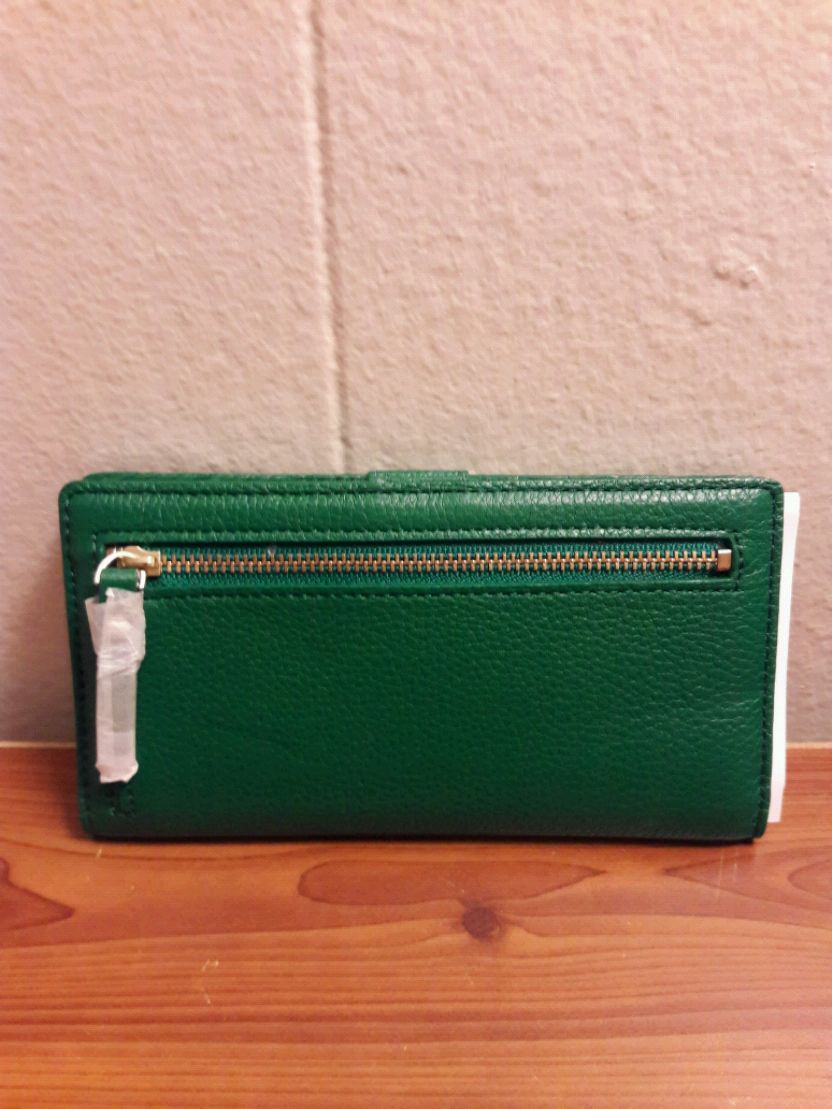}
\\

bottom\_space
& bounding box bottom to bottom of image
& \includegraphics[width=\linewidth]{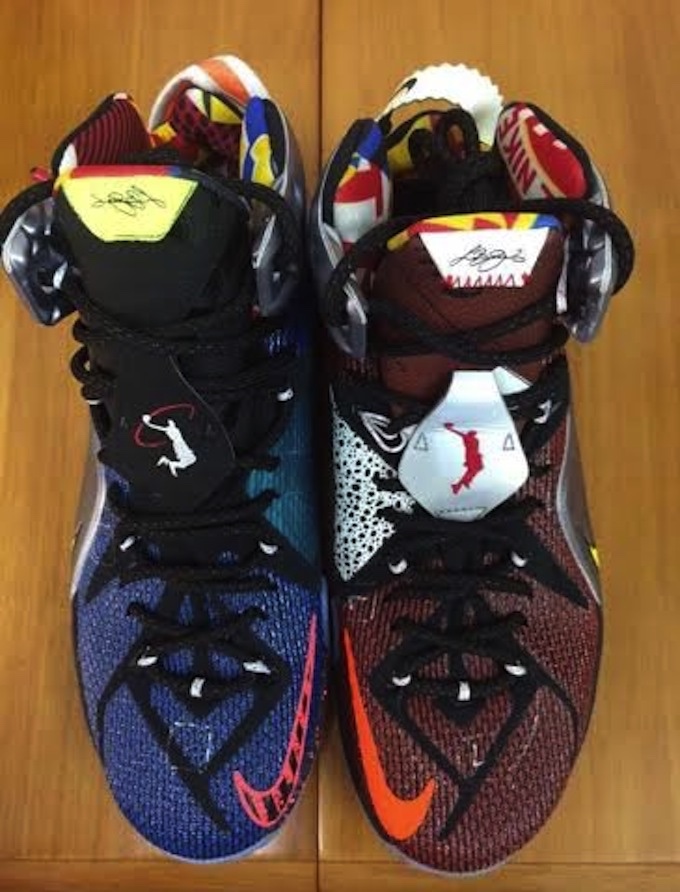}
& \includegraphics[width=\linewidth]{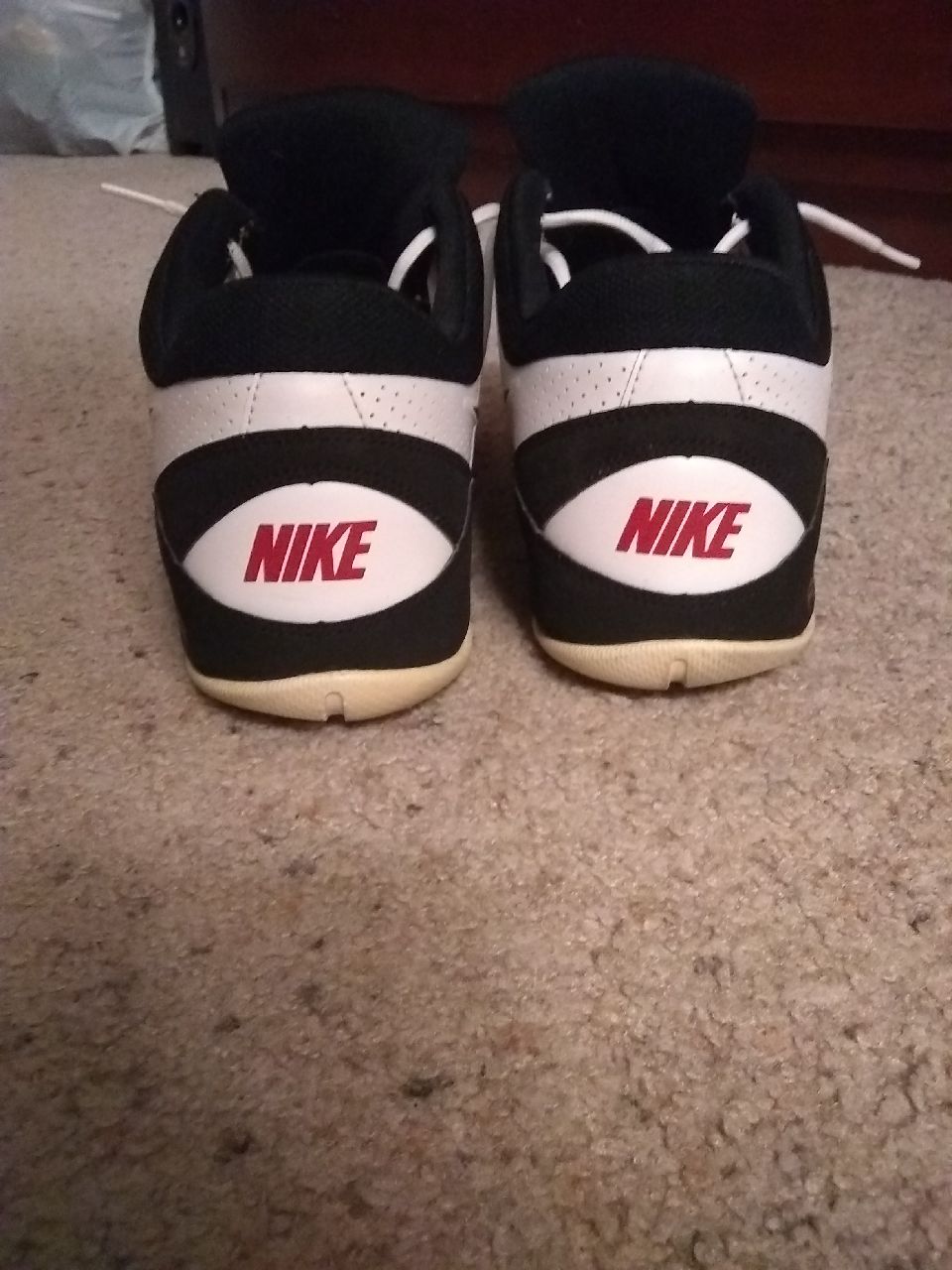}
\\

left\_space
& bounding box left to left of image
& \includegraphics[width=\linewidth]{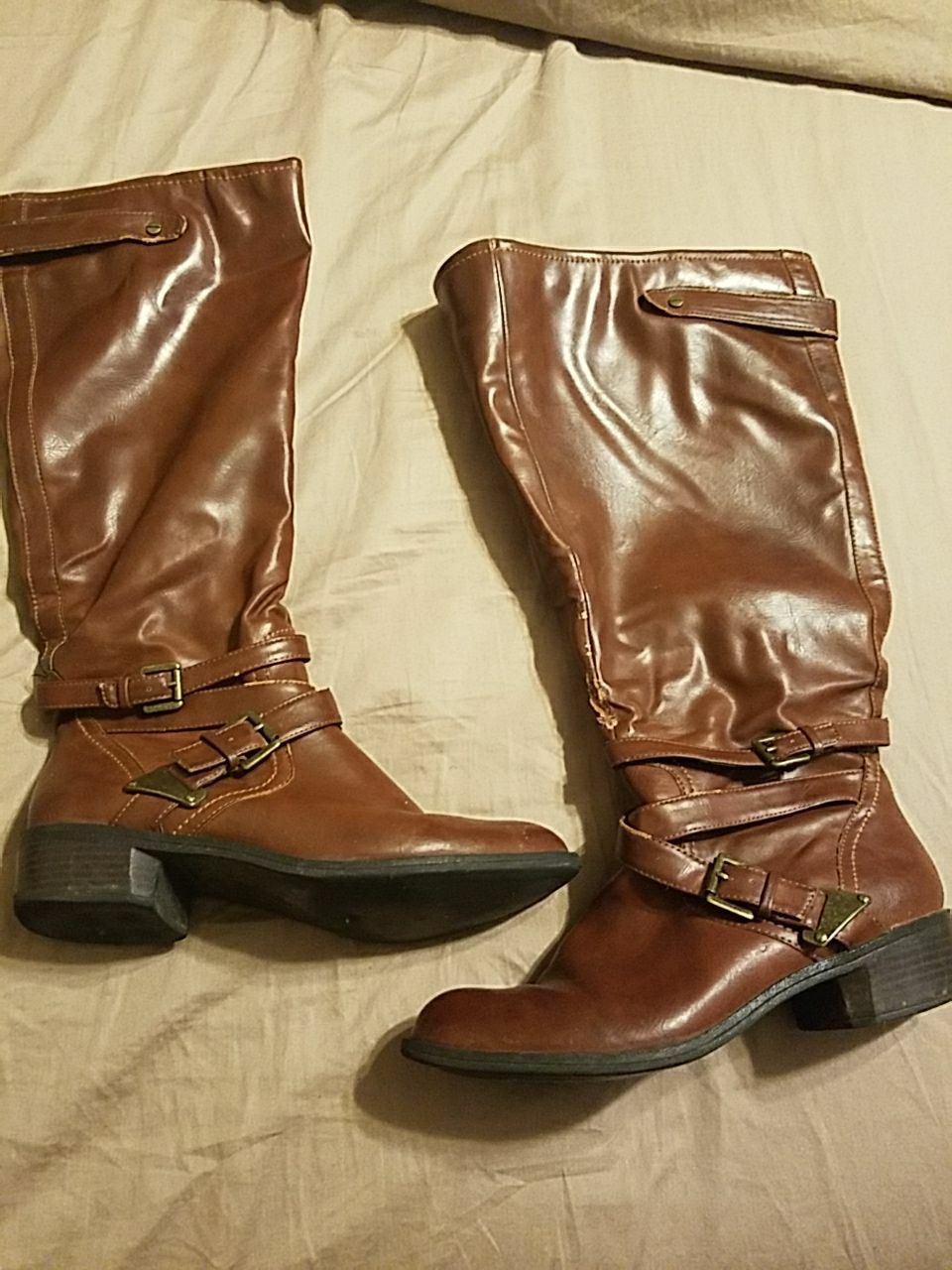}
& \includegraphics[width=\linewidth]{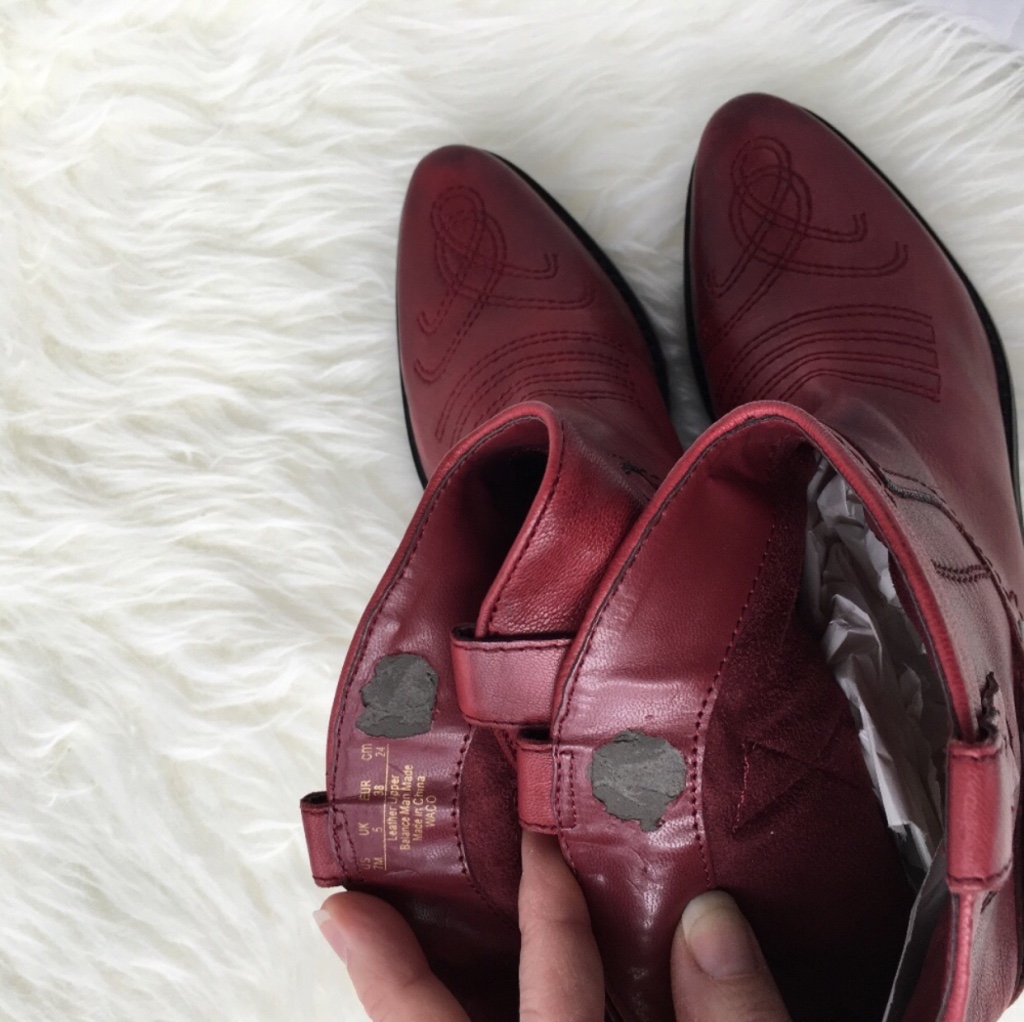}
\\

right\_space
& bounding box right to right of image
& \includegraphics[width=\linewidth]{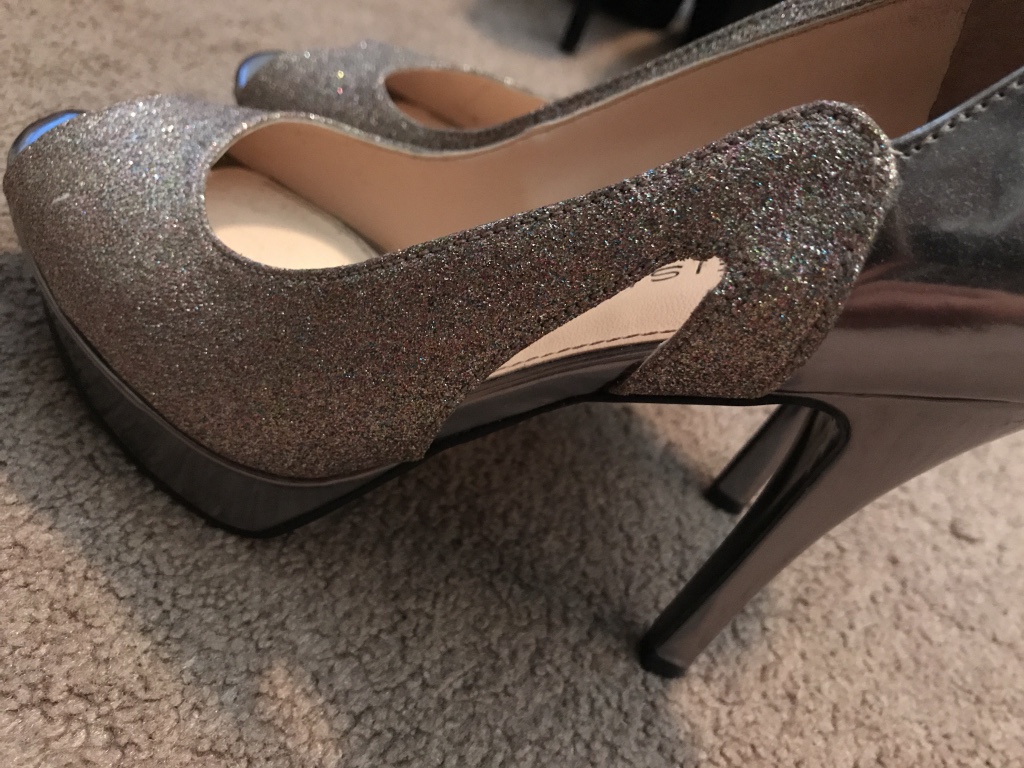}
& \includegraphics[width=\linewidth]{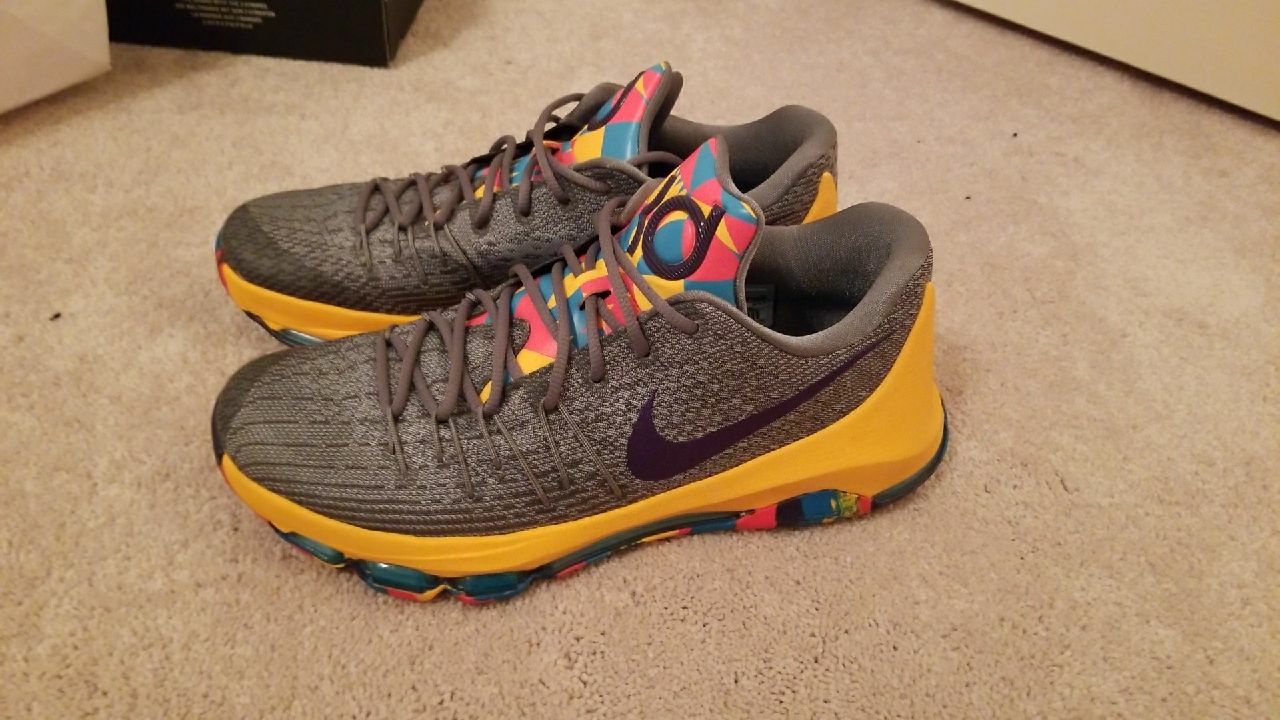}
\\

x\_asymmetry
& \multicolumn{3}{l}{abs(right\_space - left\_space)/width}
\\

y\_asymmetry
& \multicolumn{3}{l}{abs(top\_space - bottom\_space)/height}
\\

\hline
\multicolumn{4}{l}{\textbf{\textit{Regional Features:}}\textit{ (fg: foreground; bg: background)}}  \\

fgbg\_area\_ratio
& \# pixels in fg / bg
& \includegraphics[width=\linewidth]{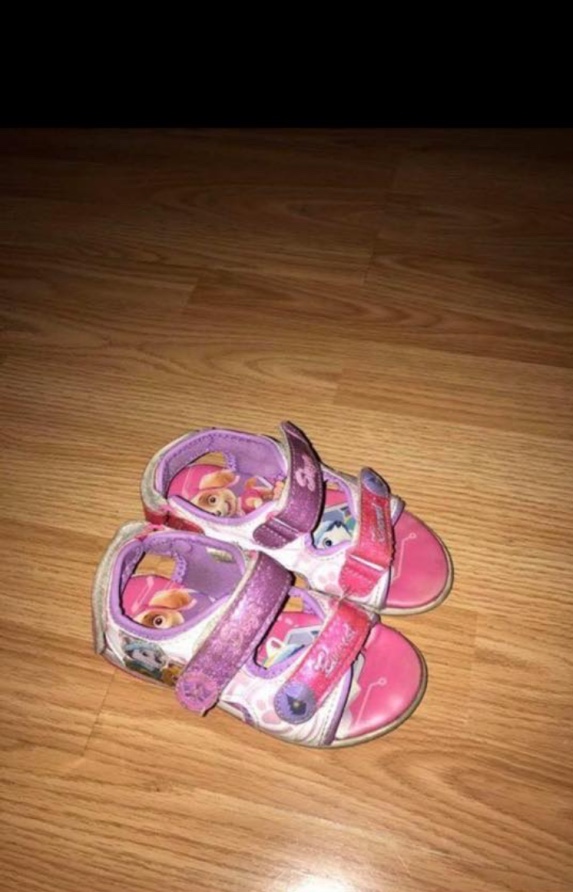}
& \includegraphics[width=\linewidth]{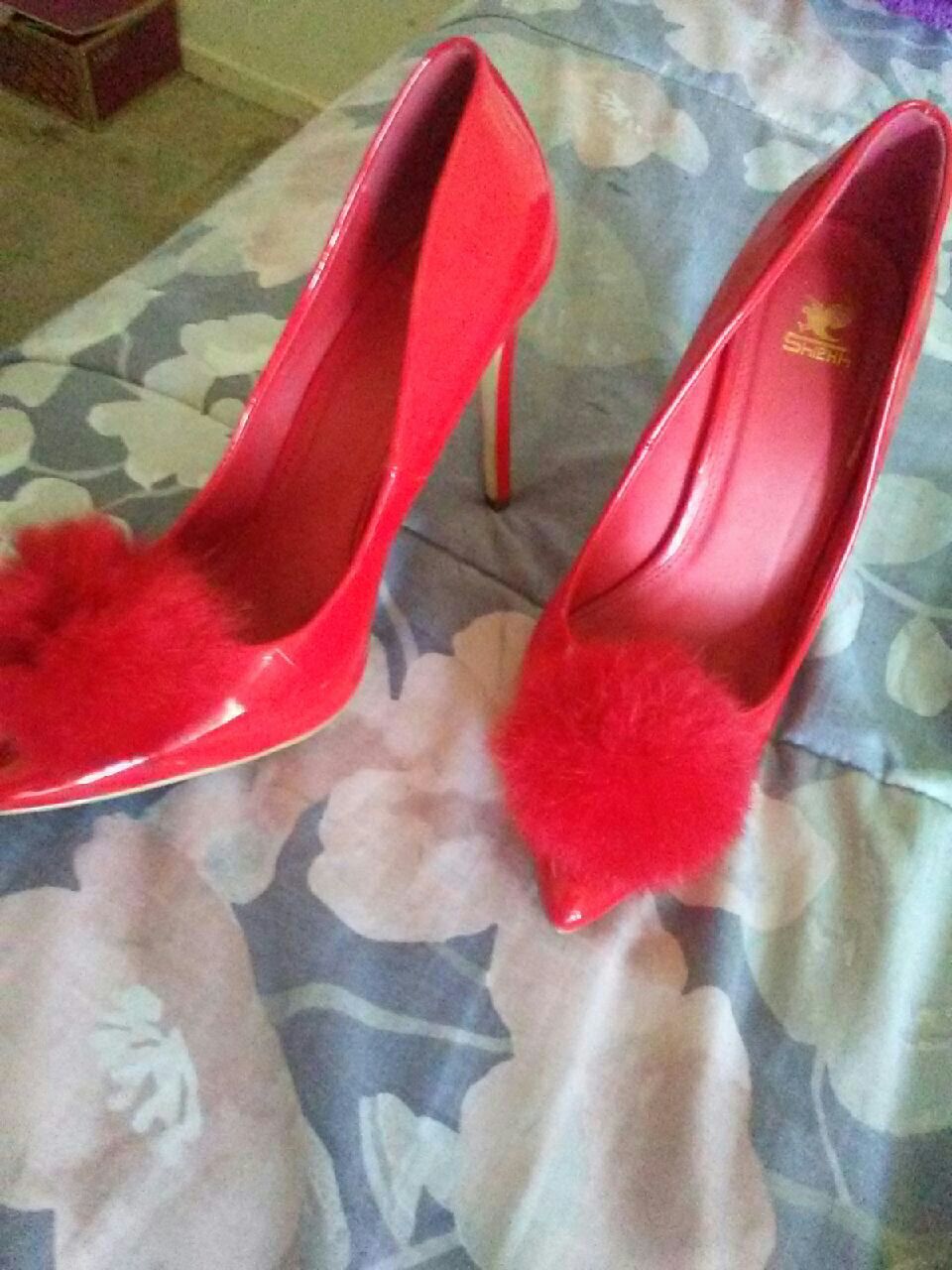}
\\

bgfg\_brightness\_diff
& brightness of bg - fg
& \includegraphics[width=\linewidth]{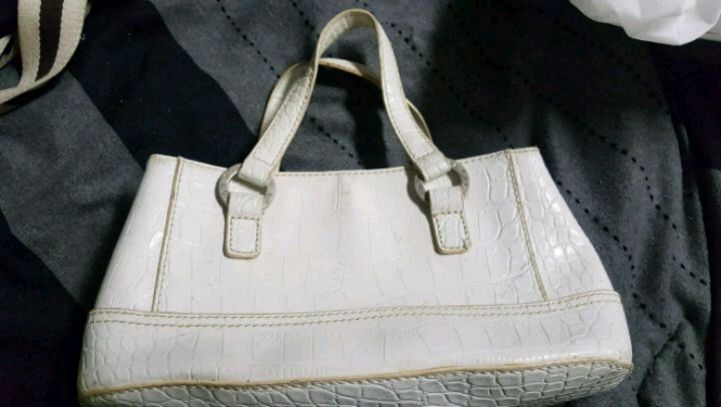}
& \includegraphics[width=\linewidth]{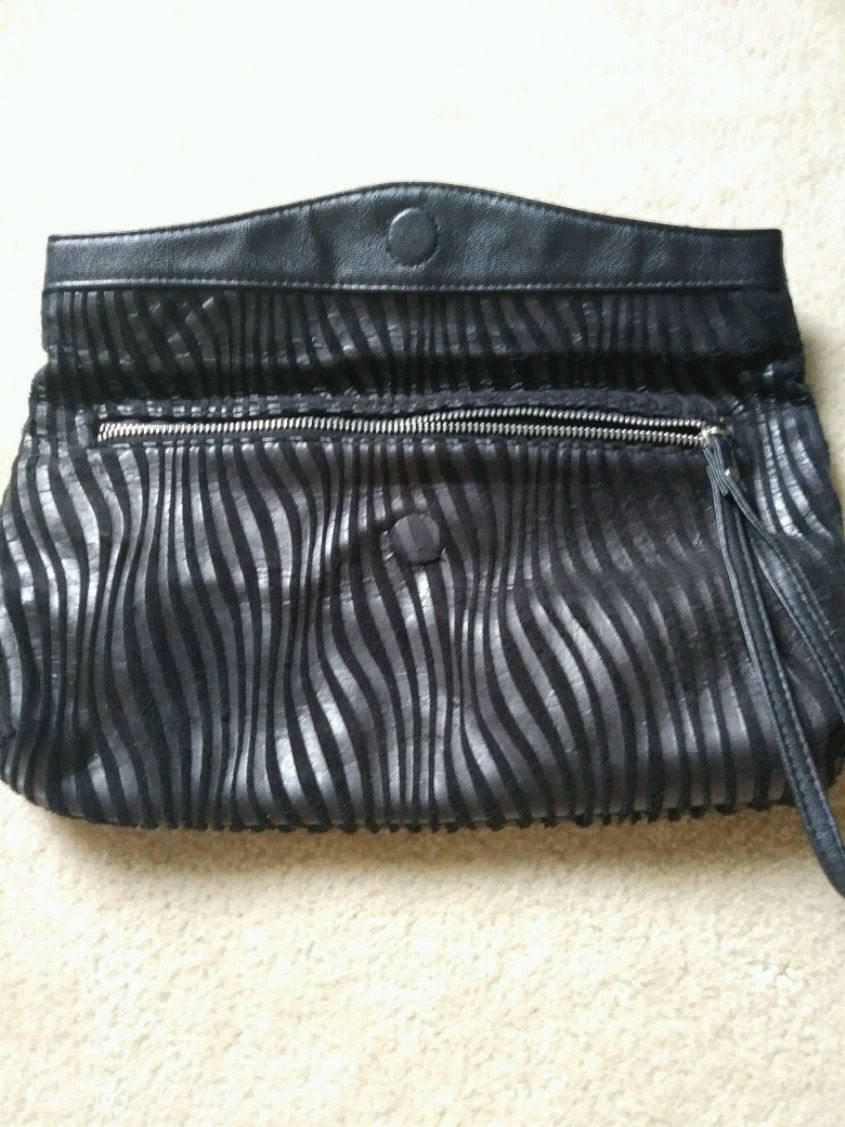}
\\

bgfg\_contrast\_diff
& contrast of bg - fg
& \includegraphics[width=\linewidth]{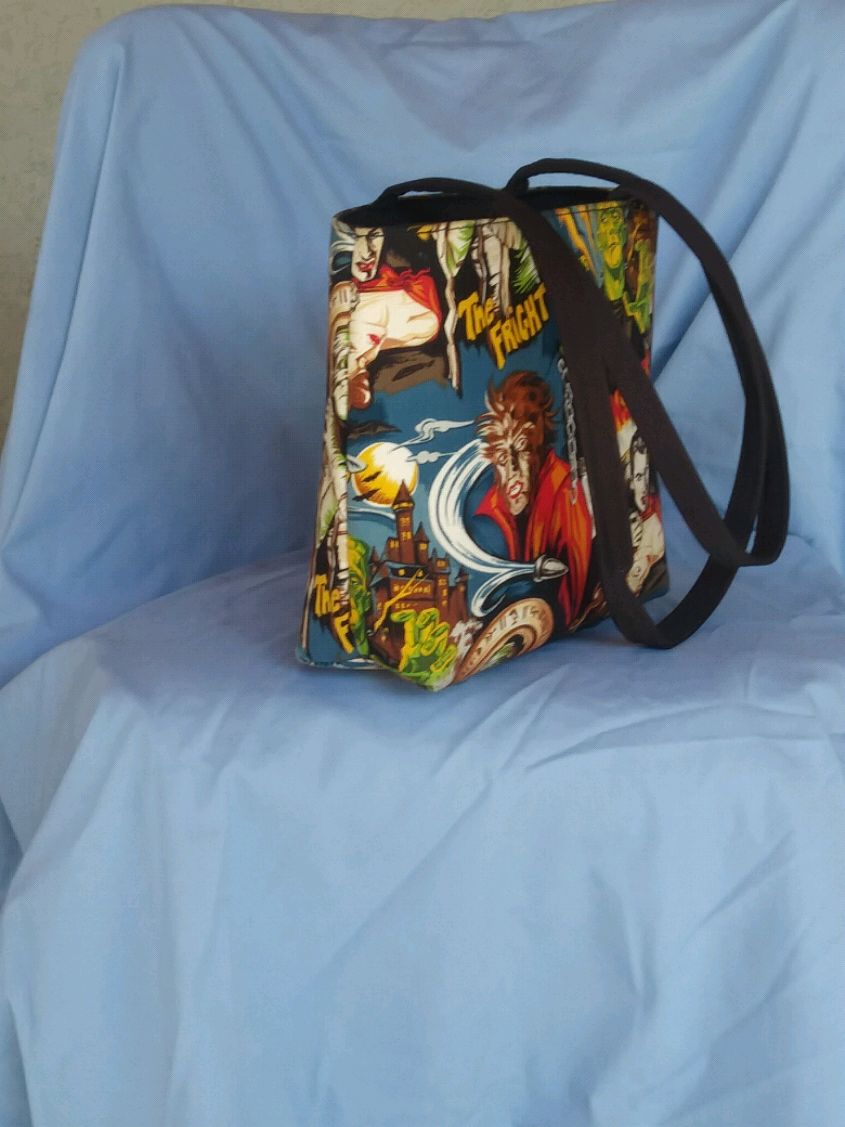}
& \includegraphics[width=\linewidth]{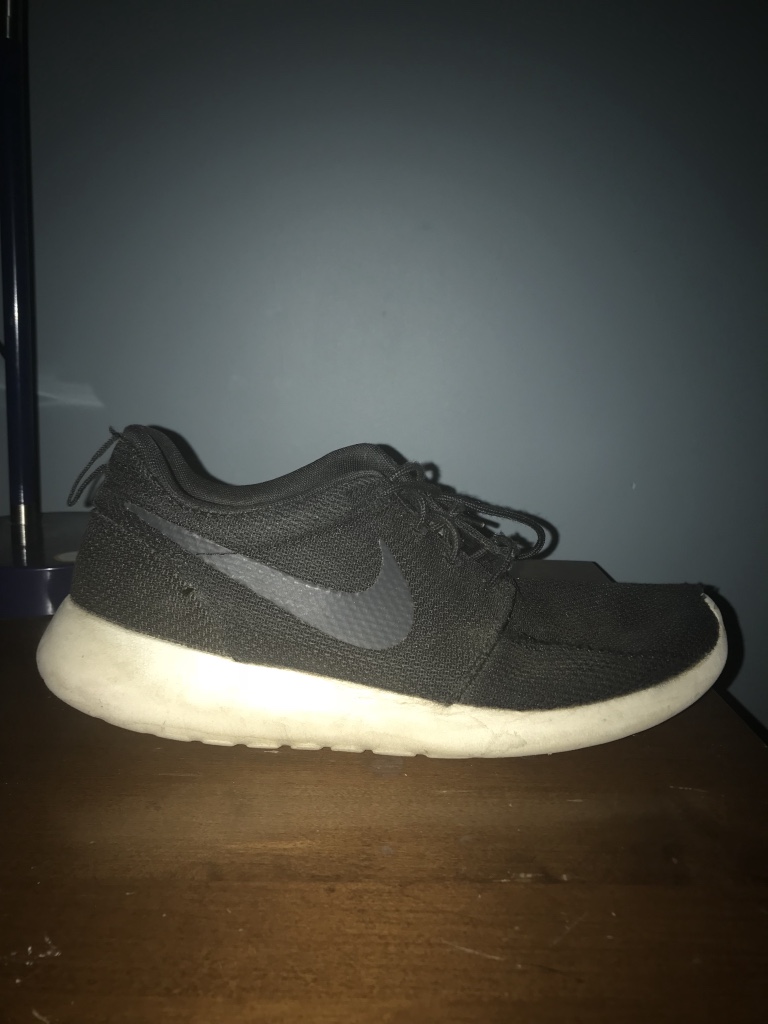}
\\

bg\_lightness
& RGB distance from a pure white image
& \includegraphics[width=\linewidth]{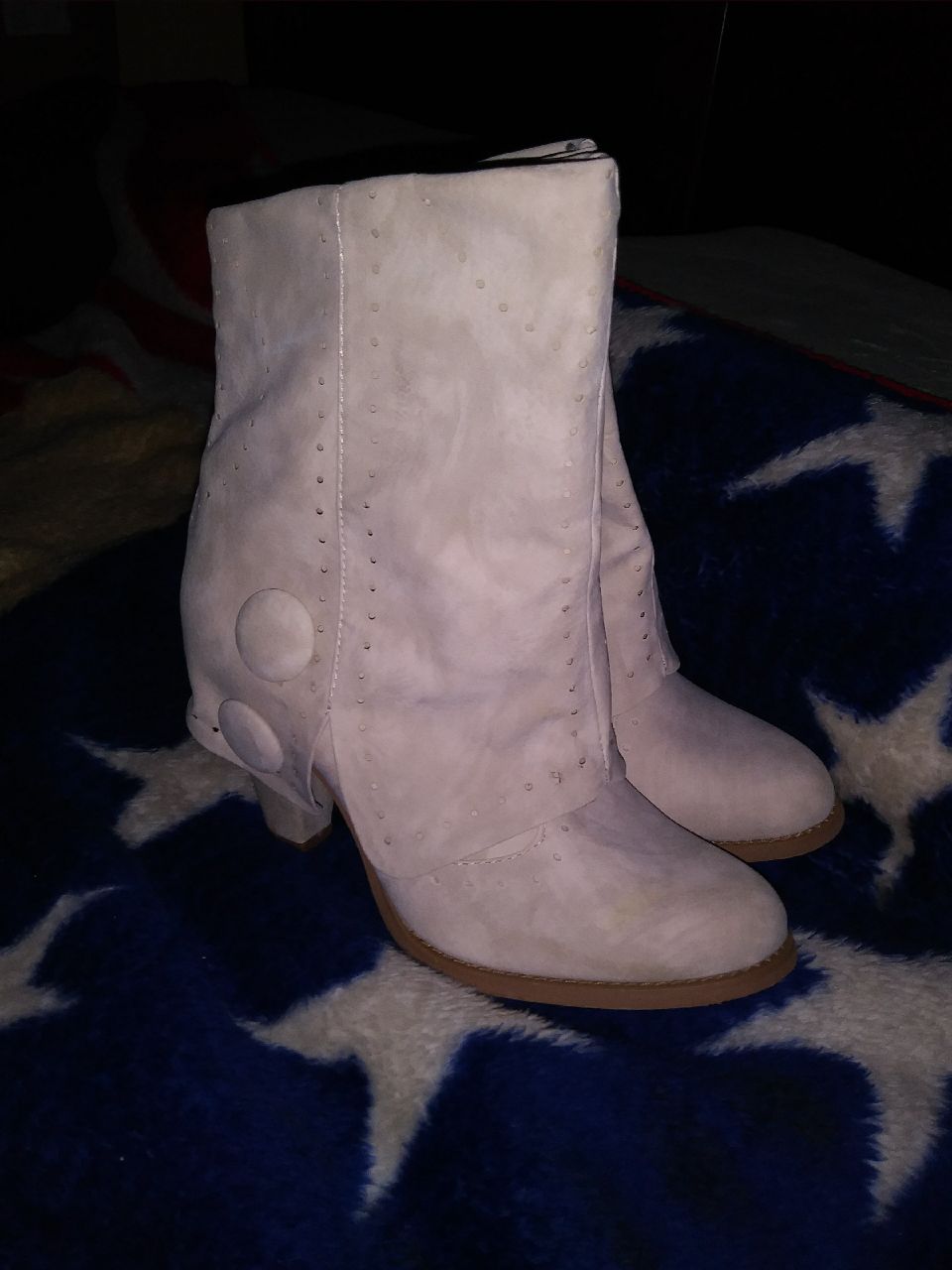}
& \includegraphics[width=\linewidth]{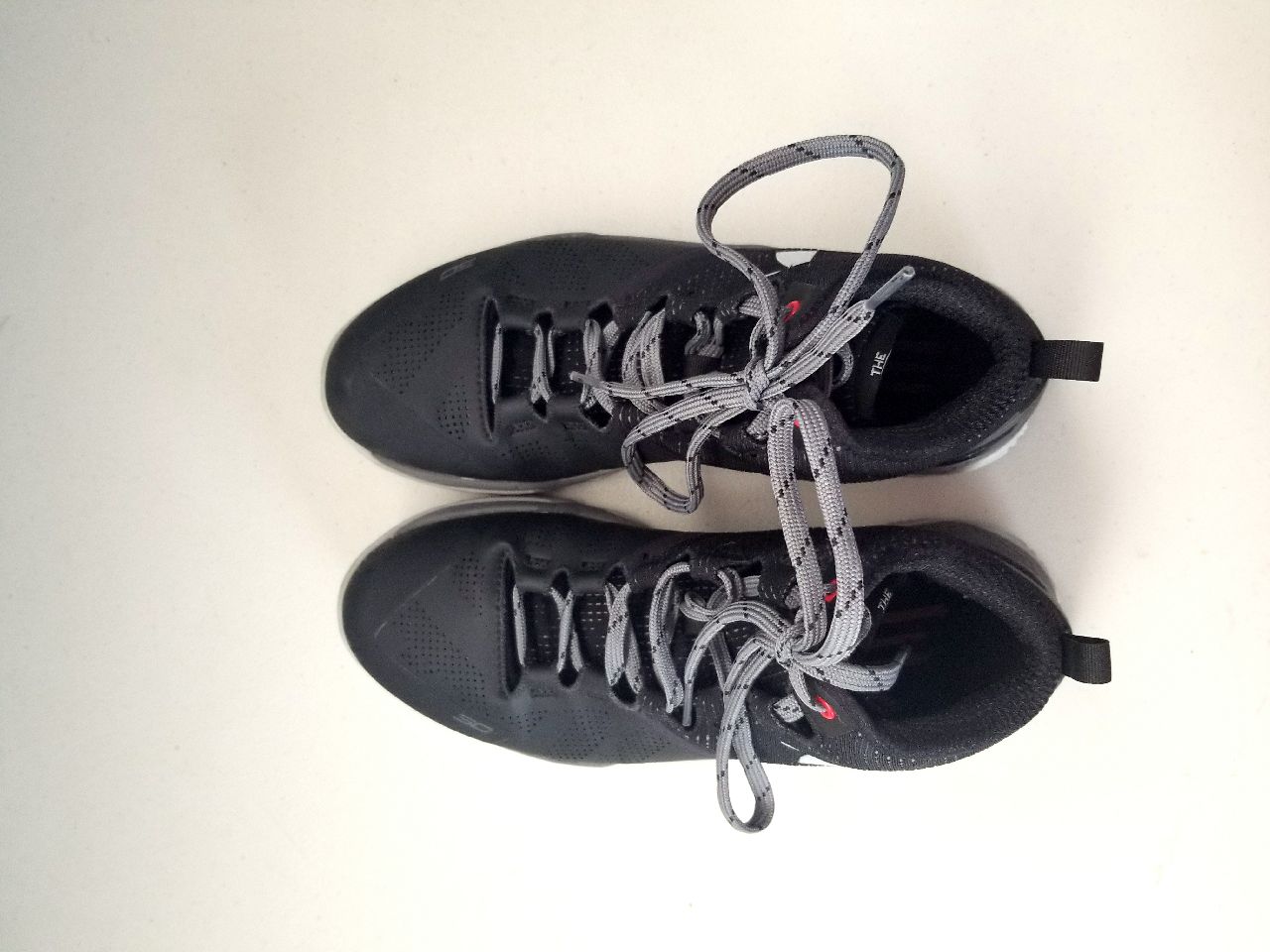}
\\

bg\_nonuniformity
& standard deviation of bg pixels in grayscale
& \includegraphics[width=\linewidth]{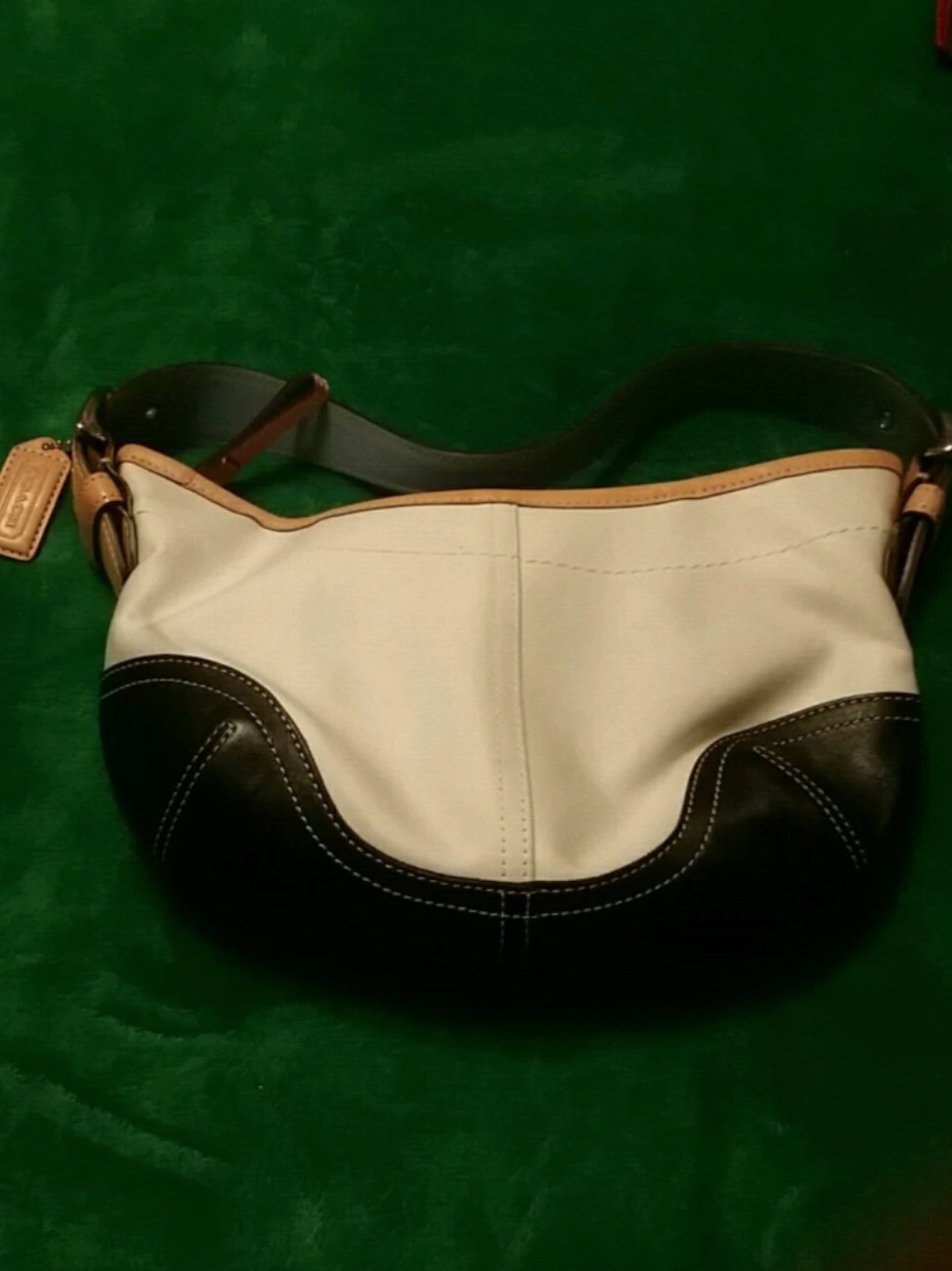}
& \includegraphics[width=\linewidth]{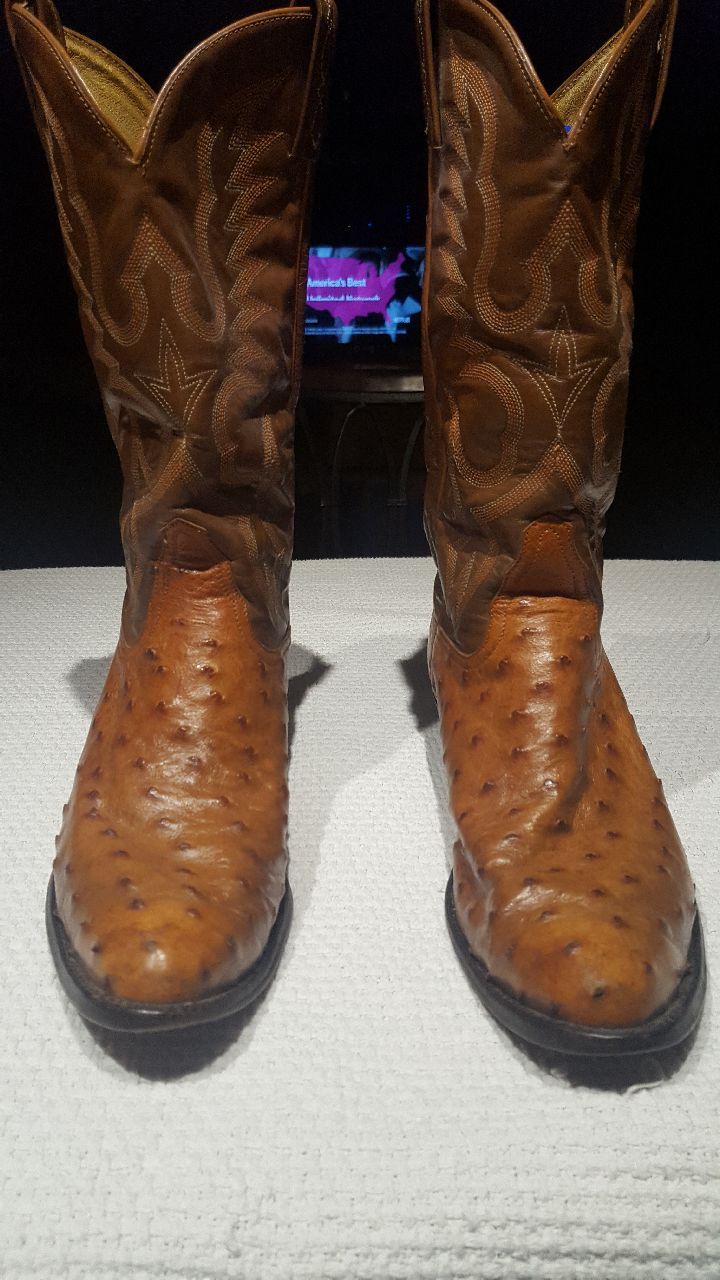}
\\

\bottomrule
\end{tabular}}
\caption{Image feature definitions and example images}
\label{tab:feature_table}
\end{table} 
\subsubsection{Calculating Image Features}

Global image features can be computed without extra information, but object and regional features require knowledge of the object that appears in the image. We trained an object detector that could detect bounding boxes for our product categories.

\textbf{Object Detection}
The process for building shoe detectors and handbag detectors is the same.
First, we collected and manually verified a dataset of 170 shoe images from the ImageNet dataset~\cite{deng2009imagenet}. Those images were already labeled with the bounding box around each shoe, and they vary across many visual styles and contexts (not just online marketplaces).
We also augmented the ImageNet images with our own from the online marketplace mentioned in Sec.~\ref{sec:online-marketplace}. We designed a crowdsourcing task and labeled 500 images from the online marketplaces image dataset. %
Crowdworkers were asked to draw the bounding box around each single shoe in the image, and each image was assigned to two distinct crowd-workers in order to ensure quality labeling. We filtered out labels where the overlap between two bounding boxes were less than 50\%. In total, we gathered 650 shoe images from both ImageNet and our online marketplace datasets, with bounding boxes around each shoe.

Next, we trained our shoe detector by using the Tensorflow Object Detection API and Single Shot Multi-box detector method~\cite{liu2016ssd}. The training set included 300 images randomly selected from our labeled datasets.
Finally, we evaluated the performance of our detector on a validation set of 350 images. We achieved a mean Average Precision (mAP0.5)
of 0.84 for the shoe detection.\footnote{mAP0.5 means an image counts as positive if the detected and groundtruth bounding boxes overlap with an intersection-over-union score greater than 0.5.} We repeated the same process for the handbag detector, and reached similar performance.

The resulted object detectors output bounding boxes around the shoes and handbags in the image, which we then used to compute object and regional features. In particular, for regional features, we used GrabCut algorithm~\cite{rother2004grabcut} to segment the foreground and background, initializing GrabCut with the detected bounding boxes as the foreground region. Then we computed the lightness and non-uniformity of the background, as well as differences in brightness and contrast between background and foreground. \autoref{tab:feature_table} contains all details and example images for our image features.

\subsubsection{Regression Analysis}

After calculating the image features, we analyzed their impact on image quality through multiple ordered logistic regression. Our dependent variable is the three image quality labels (bad, neutral, good) annotated through the crowdsourcing task in Sec.~\ref{sec:annotation}. We choose to use the image label rather than raw image quality scores for the dependent variable because there is less noise in the label (as we did majority voting to get image labels). All analysis was done on the LetGo image dataset, as that's the set of images we collected manual image quality label on.

\textbf{Detection: Shoes}
We first report on analysis of the shoe images. In our dataset, 7\% of the images did not have a target object detected, 22\% has one target object detected, 70\% had two or more targeted objects detected. A chi-squared test showed that there were significant differences in the distribution of image quality labels across different number of target objects detected ($\chi^2=463.68$, $p < .001$). Having at least one target object detected makes the image 2.7 times more likely to be labeled as Good quality.

\textbf{Detection: Handbags}
Similarly, for handbags, 8\% of our images did not have a target object detected (with the detection score threshold of $.90$). A chi-squared test showed that there was a significant difference in the distribution of image quality labels between having a target object detected and not ($\chi^2=60.34$, $p < .001$). Having the target object detected makes the image 1.4 times more likely to be labeled as Good quality.

To ensure the regression analysis remained accurate, we manually verified detection accuracy on a subset of 2,000 images of shoes and 2,000 images of handbags.
We then conducted ordered logistic regression on this manually-verified subset using image features as independent variables to predict the image quality label. Results are shown in~\autoref{tab:model_coefs}\footnote{Note the regression results do not differ significantly when we include the entire dataset, showing that features extracted from automatic object detection are robust.}.

\newcommand*{\SuperScriptSameStyle}[1]{%
  \ensuremath{%
    \mathchoice
      {{}^{\displaystyle #1}}%
      {{}^{\textstyle #1}}%
      {{}^{\scriptstyle #1}}%
      {{}^{\scriptscriptstyle #1}}%
  }%
}
\newcommand*{\oneS}{\SuperScriptSameStyle{*}}
\newcommand*{\twoS}{\SuperScriptSameStyle{**}}
\newcommand*{\threeS}{\SuperScriptSameStyle{*{*}*}}

\begin{table}[t]
{\scriptsize
\begin{tabular}{l*{4}{r@{.}l}}
\hline 
\hline \\[-1.8ex]
& \multicolumn{4}{c}{\textbf{Shoes}}
& \multicolumn{4}{c}{\textbf{Handbags}}
\\

\textbf{Feature Name}
& \multicolumn{2}{c}{Estimate}
& \multicolumn{2}{c}{SE}
& \multicolumn{2}{c}{Estimate}
& \multicolumn{2}{c}{SE} \\

\midrule

brightness
& 3&30\threeS 
& (&96)
&  3&46\threeS 
& (&92)\\

contrast
& 1&79         
& (1&26)
&  4&89\threeS 
& (1&44)\\

dynamic\_range
& 2&22\oneS   
& (1&104)
&  &47         
& (1&67)\\

resolution
& -&10         
& (&06)
& -&21         
& (&19)\\

x\_asymmetry
& -0&54         
& (&68)
& -2&26\twoS   
& (&84)\\

y\_asymmetry
& -0&74         
& (&49)
&  &73         
& (&46)\\

fgbg\_area\_ratio
& -&35\threeS 
& (&06)
& -&17\threeS 
& (&03)\\

bgfg\_brightness\_diff
& &59         
& (&48)
& -&11         
& (&53)\\

bgfg\_contrast\_diff
& -&52         
& (&54)
& -&80         
& (&42)\\

bg\_lightness
& -&46         
& (&56)
&  &16         
& (&55)\\

bg\_nonuniformity
& -1&6\oneS   
& (&633)
& -4&84\threeS 
& (&64)\\

0$\vert$1
& 3&59\twoS
& (1&20)
&  4&64\threeS 
& (1&22)\\

1$\vert$2
& 5&46\threeS
& (1&21)
&  6&13\threeS 
& (1&22)\\

\hline

AIC:
& \multicolumn{4}{c}{4170.76}
& \multicolumn{4}{c}{4169.92}
\\

\hline 
\hline \\[-1.8ex]
\end{tabular}

Significance codes:  \oneS $p < .05$ , \twoS $p < .01$ , \threeS $p < .001$
\caption{Ordered logistic regression coefficients predicting image quality labels}
\label{tab:model_coefs}
}
\end{table} 
\textbf{Brightness/Background}
On a high level, we confirmed brighter images are more likely to be labeled as high quality for both product categories. In addition, the non-uniformity of the background makes it less likely for an image to be labeled as high quality. These two features coincide with the most commonly mentioned product photography tips, background and lighting.

\textbf{Crop/Zoom}
We found mixed evidence around the crop of the images. For both product categories, higher foreground to background ratio makes it less likely for an image to be labeled as high quality, suggesting that the product should be properly framed and not too zoomed-in.

\textbf{Symmetry}
Interestingly, we found that for shoes, asymmetry does not significantly contribute to perception of quality, but for handbags, horizontal asymmetry moderately contributes to a lower perception of quality.
One potential explanation for this difference could be that shoes and handbags have different product geometry dimensions (tall v.s. wide), and sellers would take pictures with different orientations resulting in different distribution in vertical and horizontal asymmetry to begin with. Indeed, we observed that handbag images were slightly more likely to be in portrait (width$<$height) orientation than landscape orientation (24\% v.s. 22\%, $\chi^2$=$28.6$, $p$<$.001$).
Further investigation is necessary to understand how symmetry impact the perceived quality of product photos.

\textbf{Contrast}
Finally, we observed that the difference in contrast between background and foreground can impact perceived image quality. In other words, a good quality product image's background should have low contrast, and the foreground (the product) should have high contrast. The difference in brightness between background and foreground, or the lightness of the background were not significant in making an image more likely to be labeled as high quality --- suggesting a uniform background, either dark or bright, could be both effective, and potentially work better for different colored products.

\section{Marketplace Outcomes}\label{sec:purchase-outcomes}

In the previous section, we have shown that our trained models can automatically classify user-generated product photos with an accuracy of 87\% (averaged across two product categories, shoes and handbags).
Now leveraging the quality scores predicted using our models, we proceed to examine how image quality contributes to real-world and hypothetical marketplace outcomes.

We focus on two complementary marketplace outcomes:
(1) \textit{Sales}: whether an individual listing with higher quality photos is more likely to generate sales; and
(2) \textit{Perceived trustworthiness}: whether a marketplace with higher quality photos is perceived as more trustworthy.

The first outcome, sales, is naturally important for online marketplaces.
As most of these platforms operate on a fee-based model (i.e., the platform charges a flat or percentage fee when an item is sold), sales are directly linked to the revenue and success of the marketplaces.
Further, whether an item is sold is also likely associated with higher user satisfaction.

Second, the perception of whether a platform is trustworthy is important for the platform's initial adoption and growth~\cite{su10010291}.
Previous work has shown that the perceived trustworthiness of online marketplaces has a strong influence on loyalty and purchase intentions of consumers~\cite{hong2011impact}, retention~\cite{sun2010sellers}, as well as creating a price premium for the seller~\cite{pavlou2006nature}.
Several factors of the website's visual design are known to impact trust (e.g.\ complexity and colorfulness~\cite{reinecke2013predicting}), but the effect of product image quality on trustworthiness remains an open question.
\subsection{Image Quality and Sales}

For the first outcome (sales of individual listings), we used log data from eBay for analysis.
Since merchants sometimes sell many quantities of the same item, we predict whether a listing sold at least once before it expired.
To that end, a sample of balanced sold and unsold listings was created for the two product categories we studied --- shoes and handbags --- see details of data sampling in Sec.~\ref{sec:online-marketplace}.

We first used logistic regression to understand how image quality relates to trust.
We considered three different models:
(1) a baseline model using metadata information about the listings, including 
the number of days the listing has been on the platform, listing view count, and the item price;
(2) a model including image quality prediction score (the log-probability that an image will be classified as high quality by models trained on our dataset), in addition to baseline features; and
(3) a model including both image quality and aesthetic quality scores (trained on the AVA dataset), in addition to baseline features.
The results of regressions for both product categories predicting whether an item is sold are reported in~\autoref{tab:sales_prediction}.

\begin{table}
    \scriptsize
    \centering 
\begin{tabular}{@{\extracolsep{4pt}}p{.18\linewidth}*{6}{p{.07\linewidth}}}
\\[-1.8ex]\hline 
\hline \\[-1.8ex] 
\\[-1.8ex] & \multicolumn{3}{c}{Shoes ($N$=$130K$)}& \multicolumn{3}{c}{Handbags ($N$=$32K$)}\\ 
\cline{2-4} \cline{5-7} 
\\[-1.8ex] & (1) & (2) & (3) & (1) & (2) & (3)\\ 
\hline \\[-1.8ex] 
 (Intercept) & $-$1.09$^{***}$ & $-$1.14$^{***}$ & $-$1.11$^{***}$ & $-$2.66$^{***}$ & $-$2.61$^{***}$ & $-$2.42$^{***}$ \\ 
  & (0.05) & (0.05) & (0.05) & (0.07) & (0.07) & (0.07) \\ 

 \# Days (Log) & 0.26$^{***}$ & 0.26$^{***}$ & 0.26$^{***}$ & 0.64$^{***}$ & 0.63$^{***}$ & 0.60$^{***}$ \\ 
  & (0.01) & (0.01) & (0.01) & (0.01) & (0.01) & (0.01) \\ 

 \# Views (Log) & 1.07$^{***}$ & 1.08$^{***}$ & 1.08$^{***}$ & 0.94$^{***}$ & 0.95$^{***}$ & 0.98$^{***}$ \\ 
  & (0.01) & (0.01) & (0.01) & (0.01) & (0.01) & (0.01) \\
  
 Price (Log) & $-$0.57$^{***}$ & $-$0.56$^{***}$ & $-$0.57$^{***}$ & $-$0.31$^{***}$ & $-$0.32$^{***}$ & $-$0.36$^{***}$ \\ 
  & (0.01) & (0.01) & (0.01) & (0.01) & (0.01) & (0.01) \\   
 
 Img. Quality &  & 0.16$^{***}$ & 0.15$^{***}$ &  & 0.22$^{***}$ & 0.15$^{***}$ \\ 
  &  & (0.01) & (0.01) &  & (0.01) & (0.01) \\ 
 
 Img. Aesthetic &  &  & 0.08$^{***}$ &  &  & 0.31$^{***}$ \\ 
  &  &  & (0.01) &  &  & (0.02) \\ 
 
\hline \\[-1.8ex] 
AIC & 116,968 & 116,525 & 116,414 & 30,685 & 30,453 & 30,026 \\ 
\hline 
\hline \\[-1.8ex] 
\multicolumn{7}{l}{\textit{Note:} $^{*}$p$<$0.05; $^{**}$p$<$0.01; $^{***}$p$<$0.001} \\ 
\end{tabular} 
  \caption{Image quality predicted by our models is positively associated with higher likelihood that an item is sold (1.17x more for shoes, and 1.25x more for handbags).
  } 
  \label{tab:sales_prediction} 
\end{table}  

From the regression analysis, we show that image quality predicted by our models is associated with higher likelihood that an item is sold (odds ratio 1.17 for shoes, 1.25 for handbags, $p$<$.001$).
These results hold even when controlling for the predicted aesthetic quality of images.
Interestingly, both predicted image quality and aesthetic quality are more strongly associated with the sales of handbags than shoes ($\beta$=$0.22$ v.s. $0.16$ for image quality, and $\beta$=$0.31$ v.s. $0.08$ for handbags), potentially signaling that handbag is a more visual product category than shoe.

However, in terms of model performance, both image quality and aesthetic quality only resulted in small improvements to the baseline model.
We illustrate the model performance through prediction accuracy --- 10-fold cross validation showed that the baseline model can predict whether an item will be sold with an accuracy of 80.3\% for shoes, and 74.7\% for handbags.
Both image quality and aesthetic quality improved the prediction accuracy only marginally~(around 1\%).

These findings suggest that both image quality and aesthetic quality are associated with higher likelihood of sales for individual listings on online marketplaces, though both have limited power in improving the accuracy of sales prediction.
Future work could explore the difference among different product categories, or the relationship between image quality and other metrics for online marketplaces, such as ``sellability''~\cite{liu2017used}, and click-through-rate~\cite{goswami2011study,chung2012impact}.

\begin{figure*}
  \centering
  \includegraphics[width=.9\linewidth]{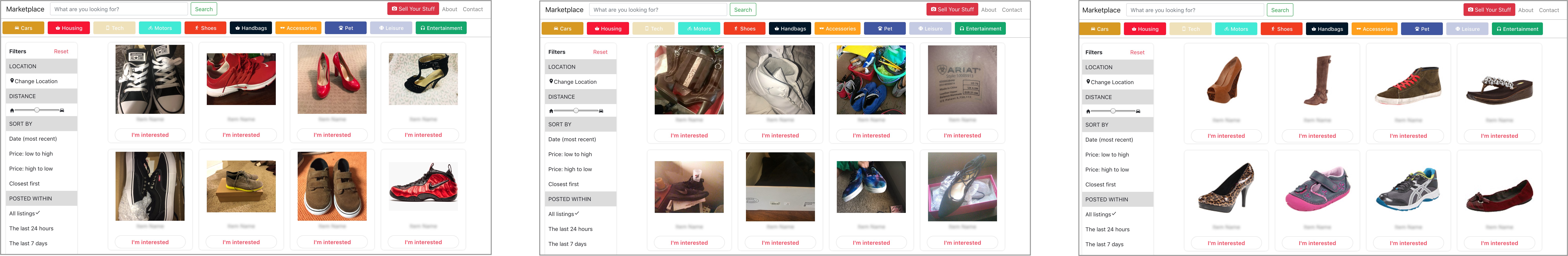}
  \caption{Hypothetical marketplace mock-ups used for our user experiment. From left to right, showing images with high quality score, low quality score, and stock imagery.}
  \label{fig:marketplace}
\end{figure*}
\subsection{Perceived Trustworthiness}\label{sec:user-trustworthiness}
For the second outcome, perceived trustworthiness of the marketplace, we designed a user experiment to compare the effects of good/bad quality user-generated images compared to stock imagery.
Our goal here is to show the potential application of our image quality models in improving the perceived trustworthiness of online marketplaces.
Our hypothesis is that high quality marketplace images (as selected by our models) will lead to the highest perception of trust, followed by stock imagery, and then low quality marketplace images will lead to the lowest perception of trust.
This hypothesis is rooted in the fact that uncertainty and information asymmetry are fundamental problems in online marketplaces that limit trust~\cite{akerlof1978market}.
Stock images, though high in aesthetic quality, do not help reduce the uncertainty of the actual conditions of the product being sold.
High quality user-generated images could help bridge the gap of information asymmetry and reduce uncertainty, therefore increasing the trust.

To test this hypothesis, we built three hypothetical marketplaces mimicking the features of popular online platforms (see~\autoref{fig:marketplace}), each populated with (1) high quality marketplace images selected by our model, (2) low quality marketplace images selected by our model, and
(3) stock imagery from the UT Zappos50K dataset~\cite{finegrained,semjitter}.
Specifically, we designed a between-subject study with three conditions, varying the images of the listings (good; bad; and stock). Each participant was randomly assigned to one condition and saw three example mock-ups of a hypothetical online peer-to-peer marketplace website, each populated with \UserExperimentUserN~images randomly drawn from a set of \UserExperimentImageCandidateN~candidate images.

We prepared the candidate images for each condition in the following ways. All of our images were from the \emph{shoe} category, but could easily be expanded to other categories.
For the ``good'' condition, the candidate images were \UserExperimentImageCandidateN~images randomly drawn from the top \UserExperimentImagePCT~LetGo images as predicted by our image quality model (to have high image quality). Similarly, candidate images used for the ``bad'' condition were drawn from the bottom \UserExperimentImagePCT~of LetGo images predicted by our image quality model.
For the ``stock'' image condition, we randomly sampled \UserExperimentImageCandidateN~images from the UT Zappos50K~\cite{finegrained,semjitter} as candidate images.

The main dependent variable for this experiment is the perceived trustworthiness of the marketplace, which could be measured in a few different aspects.
Therefore we developed a six-item trust scale based on adaptations of previous work on trust in online marketplaces~\cite{corbitt2003trust}, shown in~\autoref{tab:trust_scale}.
Each participant was requested to rate the marketplace they were shown on a 5-point Likert scale.
We take the average of participant's responses to all items in the scale as the ``trust in marketplace'' score.

\subsubsection{Results}
The experiment was pre-approved by Cornell's Institutional Review Board, under protocol \#1805007979.
We issued the task through Amazon Mechanical Turk and recruited \UserExperimentParticipantN~participants, paying 50 cents per task.

We retained 303 submissions after initial filtering, evenly distributed across three conditions.
We filtered out the submissions that were completed too fast or too slow (trimming the top and bottom 5\% based on task completion time), as previous work has shown that filtering based on completion velocity improves the quality of submissions~\cite{Ma2017SelfDisclosureAP, Wilber_2017_ICCV}.

Overall, participants reported highest level in marketplaces populated with good images, followed by stock imagery, and the lowest level of trust in marketplaces populated with bad images ($p$<$.001$).
The average perceived trustworthiness of marketplaces per condition is shown in~\autoref{fig:trust-gap}.

The ``trust gap'' between high quality user-generated images and stock imagery suggests that high ``quality'' images on online marketplaces do not necessarily have to be more aesthetically pleasing.
Our finding corroborates previous findings that show users prefer actual images over stock imagery because they give an accurate depiction of what the product looks like~\cite{bland2007risk}.
Stock images could be perceived impersonal or ``too good to be true'' in on online peer-to-peer setting.
High quality user-generated images, on the other hand, reduce uncertainty and information asymmetry in online transaction settings, therefore increasing trust.

Taken together, our marketplace experiments showed that our image quality models could effectively pick out images that are of high quality and can increase the perceived trustworthiness of online peer-to-peer marketplaces, even outperforming stock imagery.
The results of the experiment also suggest potential real-world applications of our image quality dataset as well as prediction models, by automatically ranking, filtering and selecting high quality images to present to the users to elicit feelings of trust.

\begin{table}[t]
\scriptsize
\begin{tabular}{p{.07\textwidth} p{.36\textwidth}}
\hline 
\hline \\[-1.8ex]
Item & Definition                                  \\ \hline
General
& How trustworthy do you think this marketplace is?
\\

Technical
& I believe that the chance of having a technical failure on this marketplace is quite small.
\\

Risk
& I believe that online purchases from these sellers are risky.
\\
Expectation
& I believe that products from these sellers will meet my expectations when delivered.  \\

Care
& I believe that these sellers care about its customers.
\\

Fidelity
& I believe that the photos accurately represent the condition of the products.
\\
\hline 
\hline \\[-1.8ex]
\multicolumn{2}{p{0.43\textwidth}}{\textit{Note:} Response to each item is based on a 5-point Likert scale (strongly disagree to strongly agree)}
\end{tabular}
\caption{Marketplace perceived trustworthiness scale}
\label{tab:trust_scale}

\end{table} 
\section{Conclusion}
This work attempted to develop a deeper understanding of image quality in online marketplaces through a computational approach.
By gathering and annotating a large-scale dataset of photos from online marketplaces, we were able to develop a deeper understanding of the visual factors that improve image quality, while reaching a decent accuracy ($\approx$87\%) for predicting image quality.
We have also demonstrated how predicted image quality is useful in the study of online marketplaces --- especially for trust.
High quality images selected by our model outperforms stock imagery in earning the trust of a potential buyer.

Our work is also not without limitations.
Our dataset, while large in size, only covered two product categories.
Predicted image quality also has limited prediction power in whether an item is sold.
Future work could further enrich our dataset, and explore how image quality might indirectly influence sales (e.g., through increased view count).

Our findings have important implications for the future of (increasingly mobile-based) online marketplaces; the dataset can also be useful for the broader community, e.g., providing more examples of real-world images for domain adaptation tasks.
One can also leverage the insights and data to
build tools that help merchants take better product photos and help strengthen trust between buyers and sellers.

 \ifwacvfinal%
\section{Acknowledgement}
This work is partly funded by a Facebook equipment
donation to Cornell University and by AOL through the Connected Experiences Laboratory. We additionally wish to thank our crowd workers on Mechanical Turk and our colleagues from eBay. %
\fi

\balance{}
{\small
\bibliographystyle{ieee}
\bibliography{egbib}
}

\end{document}